\documentclass{article}

    \usepackage[final]{neurips_2021}

\usepackage[utf8]{inputenc} 
\usepackage[T1]{fontenc}    
\PassOptionsToPackage{hyphens}{url}\usepackage{hyperref}
\usepackage[hyphens]{url}   
\usepackage{booktabs}       
\usepackage{amsfonts}       
\usepackage{nicefrac}       
\usepackage{microtype}      
\usepackage{xcolor}         

\newcommand{\includeAppendix}{yes}

\usepackage{times}

\usepackage{soul}

\usepackage{algorithm}
\usepackage{algorithmic}
\usepackage{amsmath}
\usepackage{amssymb}
\usepackage{amsthm}
\usepackage[small]{caption}
\usepackage{centernot}
\usepackage{color}
\usepackage{dsfont}
\usepackage{enumitem}
\usepackage{eqnarray}
\usepackage{graphicx}
\usepackage{ifthen}
\usepackage{listings}
\usepackage{nccmath}
\usepackage{mathtools}
\usepackage{outlines}
\usepackage{textcomp}[texttildelow]
\usepackage{relsize}
\usepackage{sistyle}
\SIthousandsep{,}
\usepackage{subcaption}
\usepackage{stackengine}
\usepackage{thmtools}
\usepackage[normalem]{ulem}
\usepackage{xcolor}
\usepackage{centernot}
\usepackage[makeroom]{cancel}

\usepackage{etoc}

\hypersetup{
    colorlinks,
    linkcolor={black!50!black},
    citecolor={blue!50!black},
    urlcolor={blue!50!black}
}

\newcommand{\REF}{\textcolor{red}{REF}}

\newcommand{\appref}[2]{\ifthenelse{\equal{\includeAppendix}{yes}}{\ref{#1}}{\REF}}

\DeclareMathOperator{\E}{\mathbb{E}}

\newcommand\ind{\mathds{1}}
\newcommand\taghere{\addtocounter{equation}{1}\tag{\theequation}}

\declaretheoremstyle[
	spaceabove=12pt,
	bodyfont=\itshape,
]{plain}
\theoremstyle{plain}
\declaretheorem{theorem}
\declaretheorem[numberwithin=section]{lemma}
\declaretheorem[parent=theorem]{corollary}

\declaretheoremstyle[
	spaceabove=6pt,
	headfont=\normalfont\bfseries,
	notefont=\normalfont\bfseries,
	notebraces={}{},
	bodyfont=\normalfont,
	qed=\qedsymbol,
	headpunct={:},
]{proofstyle}
\declaretheorem[unnumbered, style=proofstyle]{Proof}

\declaretheoremstyle[
	spaceabove=6pt,
	notefont=\bfseries,
	notebraces={(}{)},
	bodyfont=\itshape,
]{defnbold}
\declaretheorem[style=defnbold]{definition}

\title{Learning Markov State Abstractions\\for Deep Reinforcement Learning}

\author{%
  Cameron Allen\thanks{Please send any correspondence to Cameron Allen $<$\href{mailto:csal@brown.edu}{\texttt{csal@brown.edu}}$>$. Code repository available at \url{https://github.com/camall3n/markov-state-abstractions}.} \\
  Brown University\\
  \And
  Neev Parikh \\
  Brown University\\
  \And
  Omer Gottesman \\
  Brown University\\
  \And
  George Konidaris \\
  Brown University\\
}

\begin{document}

\maketitle

\DeclareRobustCommand{\VAN}[3]{#2} 


\begin{abstract}
  A fundamental assumption of reinforcement learning in Markov decision processes (MDPs) is that the relevant decision process is, in fact, Markov. However, when MDPs have rich observations, agents typically learn by way of an abstract state representation, and such representations are not guaranteed to preserve the Markov property. We introduce a novel set of conditions and prove that they are sufficient for learning a Markov abstract state representation. We then describe a practical training procedure that combines inverse model estimation and temporal contrastive learning to learn an abstraction that approximately satisfies these conditions. Our novel training objective is compatible with both online and offline training: it does not require a reward signal, but agents can capitalize on reward information when available. We empirically evaluate our approach on a visual gridworld domain and a set of continuous control benchmarks. Our approach learns representations that capture the underlying structure of the domain and lead to improved sample efficiency over state-of-the-art deep reinforcement learning with visual features---often matching or exceeding the performance achieved with hand-designed compact state information.
\end{abstract}

\section{Introduction}
Reinforcement learning (RL) in Markov decision processes with rich observations requires a suitable state representation.
Typically, such representations are learned implicitly as a byproduct of doing deep RL.
However, in domains where precise and succinct expert state information is available, agents trained on such expert state features usually outperform agents trained on rich observations.
Much recent work \citep{shelhamer2016loss,pathak2017curiosity,ha2018world,gelada2019deepmdp,yarats2019improving,kaiser2019model,laskin2020reinforcement,laskin2020curl,zhang2021learning} has sought to close this \emph{representation gap} by incorporating a wide range of representation-learning objectives that help the agent learn abstract representations with various desirable properties.


\begin{figure}[t]
    \centering
    \includegraphics[width=\columnwidth]{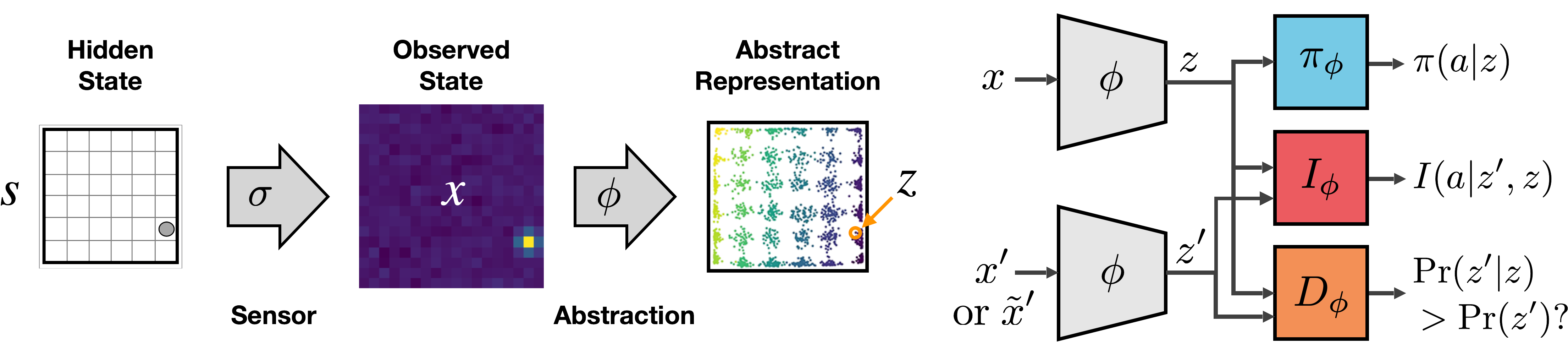}
\caption{(Left) A $6\times 6$ visual gridworld domain with hidden state $s$ and unknown sensor $\sigma$, where an abstraction function $\phi$ maps each high-dimensional observed state $x$ to a lower-dimensional abstract state $z$ (orange circle). (Right) Our Markov abstraction training architecture. A shared encoder $\phi$ maps ground states $x,x'$ to abstract states $z,z'$, which are inputs to an inverse dynamics model $I$ and a contrastive model $D$ that discriminates between real and fake state transitions. The agent's policy $\pi$ depends only on the current abstract state.}
\label{fig:main-objective}
\end{figure}

Perhaps the most obvious property to incentivize in a state representation is the Markov property, which holds if and only if the representation contains enough information to accurately characterize the rewards and transition dynamics of the decision process.
Markov decision processes (MDPs) have this property by definition, and most reinforcement learning algorithms depend on having Markov state representations.
For instance, the ubiquitous objective of learning a stationary, state-dependent optimal policy that specifies how the agent should behave is \emph{only} appropriate for \emph{Markov} states.


But learned abstract state representations are not necessarily Markov, even when built on top of MDPs.
This is due to the fact that abstraction necessarily throws away information.
Discard too much information, and the resulting representation cannot accurately characterize the environment.
Discard too little, and agents will fail to close the representation gap.
Abstraction must balance between ignoring irrelevant information and preserving what is important for decision making.
If reward feedback is available, an agent can use it to determine which state information
is relevant to the task at hand.
Alternatively, if the agent can predict raw environment observations from learned abstract states, then \emph{all} available information is preserved (along with the Markov property).
However, these approaches are impractical when rewards are sparse or non-existent, or observations are sufficiently complex.

We introduce a new approach to learning Markov state abstractions.
We begin by defining a set of theoretical conditions that are sufficient for an abstraction to retain the Markov property.
We next show that these conditions are approximately satisfied by simultaneously training an inverse model to predict the action distribution that explains two consecutive states, and a discriminator to determine whether two given states were in fact consecutive.
Our combined training objective (architecture shown in Fig. \ref{fig:main-objective}, right) supports learning Markov abstract representations without requiring reward information or observation prediction.

Our method is effective for learning Markov state abstractions that are highly beneficial for decision making.
We perform evaluations in two settings with rich (visual) observations: a gridworld navigation task (Fig. \ref{fig:main-objective}, left) and a set of continuous control benchmarks.
In the gridworld, we construct an abstract representation offline---without access to reward feedback---that captures the underlying structure of the domain and fully closes the representation gap between visual and expert features.
In the control benchmarks, we combine our training objective (online) with traditional RL, where it leads to a significant performance improvement over state-of-the-art visual representation learning.


\section{Background}
\label{sec:background}
A Markov decision process $M$ consists of sets of states $X$ and actions $A$, reward function $R: X \times A \times X \rightarrow \mathbb{R}$, transition dynamics $T: X\times A \rightarrow \Pr(X)$, and discount factor $\gamma$.
For our theoretical results, we assume $M$ is a Block MDP \citep{du2019provably} whose behavior is governed by a much smaller (but unobserved) set of states $S$, and where $X$ is a rich observation generated by a noisy sensor function $\sigma:S\rightarrow \Pr(X)$, as in Figure \ref{fig:main-objective} (left).
Block MDPs conveniently avoid potential issues arising from partial observability by assuming that each observation uniquely identifies the unobserved state that generated it.
In other words, there exists a perfect ``inverse sensor'' function $\sigma^{-1}(x) \mapsto s$, which means the observations are themselves Markov, as we define below.
Note that $S$, $\sigma$, and $\sigma^{-1}$ are all unknown to the agent.

\begin{definition}[Markov State Representation]
\label{def:ground-markov}
A decision process $M = (X, A, R, T, \gamma)$ and its state representation $X$ are \emph{Markov} if and only if $T^{(k)}\left(x_{t+1}|\{a_{t-i},x_{t-i}\}_{i=0}^{k}\right) = T(x_{t+1}|a_t,x_t)$ and $R^{(k)}\left(x_{t+1},\{a_{t-i},x_{t-i}\}_{i=0}^{k}\right) = R(x_{t+1},a_t,x_t)$, for all $a \in A$, $x \in X$, $k \ge 1$.
\end{definition}
The superscript $(k)$ denotes that the function is being conditioned on $k$ additional steps of history.

The Markov property means that each state $X$ it is a sufficient statistic for predicting the next state and expected reward, for any action the agent might select.
Technically, each state must also be sufficient for determining the set of actions available to the agent in that state, but here we assume, as is common, that every action is available in every state.

The behavior of an RL agent is typically determined by a (Markov) policy $\pi: X \rightarrow \Pr(A)$, and each policy induces value function $V^\pi: X \rightarrow \mathbb{R}$, which is defined as the expected sum of future discounted rewards starting from a given state and following the policy $\pi$ thereafter.
The agent's objective is to learn an optimal policy $\pi^*$ that maximizes value at every state.
Note that the assumption that the optimal policy is stationary and Markov---that it only depends on state---is appropriate only if the decision process itself is Markov; almost all RL algorithms simply assume this to be true.


\subsection{State Abstraction}
\label{sec:bg-state-abstraction}
To support decision making when $X$ is too high-dimensional for tractable learning, we turn to state abstraction.
Our objective is to find an abstraction function $\phi:X \rightarrow Z$ mapping each ground state\footnote{We refer to $x\in X$ as \emph{ground states} (and $M$ as the \emph{ground MDP}), to reflect that these quantities are \emph{grounded}, as opposed to abstract, i.e. they have a firm basis in the true environment.} $x$ to an abstract state $z=\phi(x)$, with the hope that learning is tractable using the abstract representation $Z$ (see Figure \ref{fig:main-objective}, left).
Since our goal is to support effective abstract decision making, we are mainly concerned with the policy class $\Pi_\phi$, the set of policies with the same behavior for all ground states that share the same abstract state:
\[\Pi_\phi := \left\{\pi:\big(\phi(x_1)=\phi(x_2)\big) \implies \big(\pi(a|x_1)=\pi(a|x_2)\big),\ \forall\ a\in A; x_1,x_2 \in X \right\}. \taghere \label{eqn:abstract-policy}\]

An abstraction $\phi: X\rightarrow Z$, when applied to an MDP $M$, induces a new abstract decision process $M_\phi = (Z, A, T_{\phi,t}^\pi, R_{\phi,t}^\pi, \gamma)$, whose dynamics may depend on the current timestep $t$ or the agent's behavior policy $\pi$, and, crucially, which is not necessarily Markov.
\begin{figure}[!h]
  \centering
  \includegraphics[width=.6\textwidth]{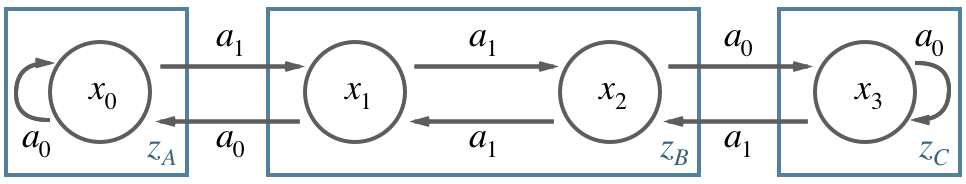}
  \captionof{figure}{An MDP and a non-Markov abstraction.}
  \label{fig:inverse-counterexample}
\end{figure}
Consider the following example. The figure above depicts a four-state, two-action MDP, and an abstraction $\phi$ where $\phi(x_1)=\phi(x_2)=z_B$. It is common to view state abstraction as aggregating or partitioning ground states into abstract states in this way \citep{li2006towards}. Generally this involves choosing a fixed weighting scheme $w(x)$ to express how much each ground state $x$ contributes to its abstract state $z$, where the weights sum to 1 for the set of $x$ in each $z$. We can then define the abstract transition dynamics $T_\phi$ as a $w$-weighted sum of the ground dynamics (and similarly for reward): $T_\phi(z'|a,z) = \sum_{x'\in z'} \sum_{x\in z} T(x'|a,x) w(x)$. A natural choice for $w(x)$ is to use the ground-state visitation frequencies. For example, if the agent selects actions uniformly at random, this leads to $w(x_0)=w(x_3)=1$ and $w(x_1)=w(x_2)=0.5$.

In this formulation, the abstract decision process is assumed to be Markov by construction, and $T_\phi$ and $R_\phi$ are assumed not to depend on the policy or timestep.
But this is an oversimplification.
The abstract transition dynamics are \emph{not} Markov; they change depending on how much history is conditioned on: $T_\phi(z_A|a_t=a_0,z_t=z_B) = 0.5$, whereas $\Pr(z_A|a_t=a_0,z_t=z_B,a_{t-1}=a_1,z_{t-1}=z_A) = 1$.
By contrast, the ground MDP's dynamics are fully deterministic (and Markov).
Clearly, if we define the abstract MDP in this way, it may not match the behavior of the original MDP.\footnote{\citet{abel2018salrl} presented a three-state chain MDP where they made a similar observation.}
Even worse, if the agent's policy changes---such as during learning---this discrepancy can cause RL algorithms with bounded sample complexity in the ground MDP to make an arbitrarily large number of mistakes in the abstract MDP \citep{abel2018salrl}.

For the abstract decision process to faithfully simulate the ground MDP's dynamics, $w(x)$ must be allowed to vary such that it always reflects the correct ground-state frequencies.
Unfortunately, even in the simple example above, maintaining accurate weights for $w(x)$ requires keeping track of an unbounded amount of history: if an agent reaches abstract state $z_B$ and repeats action $a_1$ an arbitrarily large number of times ($N$), then knowing precisely which ground state it will end up in ($x_1$ or $x_2$) requires remembering the abstract state it originally came from $N+1$ steps prior.
Successful modeling of the transition dynamics for a subsequent action $a_0$ hinges on exactly this distinction between $x_1$ and $x_2$. The abstraction introduces partial observability, and to compensate, $w(x)$ must be replaced with a belief distribution over ground states, conditioned on the entire history; instead of an abstract MDP, we have an abstract POMDP \citep{bai2016markovian}.
This is especially unsatisfying, not just because learning is challenging in POMDPs \citep{zhang2012covering}, but because formulating the problem as a Block MDP was supposed to avoid precisely this type of partial observability: an abstraction exists (namely $\phi = \sigma^{-1}$) that would result in a fully observable abstract MDP, if only the agent knew what it was.

We show that it is possible to learn a state abstraction that both reflects the behavior of the underlying ground MDP \emph{and} preserves the Markov property in the abstract MDP, without ever estimating or maintaining a belief distribution. Our theoretical result leverages inverse dynamics models and contrastive learning to derive conditions under which a state abstraction is Markov. We then adapt these conditions into a corresponding training objective for learning such an abstraction directly from the agent's experience in the ground MDP.

\section{Related Work}
\label{sec:related-work}
\subsection{Bisimulation}
The idea of learning Markov state abstractions is related to the concept of bisimulation \citep{dean1997model}, the strictest type of state aggregation discussed in \citet{li2006towards}, where ground states are equivalent if they have exactly the same expected reward and transition dynamics.
Preserving the Markov property is a prerequisite for a bisimulation abstraction, since the abstraction must also preserve the (Markov) ground-state transition dynamics.
Bisimulation-based abstraction is appealing because, by definition, it leads to high-fidelity representations. But bisimulation is also very restrictive, because it requires $\phi$ to be Markov for \emph{any} policy (rather than just those in $\Pi_\phi$).

Subsequent work on approximate MDP homomorphisms \citep{ravindran2004approximate} and bisimulation metrics \citep{ferns2004metrics,ferns2011bisimulation} relaxed these strict assumptions and
allowed ground states to have varying degrees of ``bisimilarity.''
\citet{castro2020scalable} introduced a further relaxation, $\pi$-bisimulation, which measures the behavioral similarity of states \emph{under a policy $\pi$}. But whereas full bisimulation can be too strong, since it constrains the representation based on policies the agent may never actually select, $\pi$-bisimulation can be too weak, since if the policy deviates from $\pi$ (e.g. during learning), the metric must be updated, and the representation along with it. Our approach can be thought of as a useful compromise between these two extremes.

While bisimulation-based approaches have historically been computationally expensive and difficult to scale, recent work has started to change that \citep{castro2020scalable,lehnert2020successor,van_der_pol2020plannable,biza2021learning}.
Two recent algorithms in particular, DeepMDP \citep{gelada2019deepmdp} and Deep Bisimulation for Control (DBC) \citep{zhang2021learning}, learn approximate bisimulation abstractions by training the abstraction end-to-end with an abstract transition model and reward function. This is a rather straightforward way to learn Markov abstract state representations since it effectively encodes Definition \ref{def:ground-markov} as a loss function.

One drawback of bisimulation-based methods is that learning an accurate model can be challenging and typically requires restrictive modeling assumptions, such as deterministic, linear, or Gaussian transition dynamics.
Bisimulation methods may also struggle if rewards are sparse or if the abstraction must be learned without access to rewards.
Jointly training an abstraction $\phi$ with only the transition model $\widehat{T}(\phi(x),a)\approx \phi(x')$ can easily lead to a trivial abstraction like $\phi(x)\mapsto 0$ for all $x$, since $\phi$ produces both the inputs and outputs for the model.
Our approach to learning a Markov abstraction avoids this type of representation collapse without learning a forward model, and is less restrictive than bisimulation, since it is compatible with both reward-free and reward-informed settings.

\subsection{Ground-State Prediction and Reconstruction}
\label{sec:bg-pixel-prediction}
Ground-state (or pixel) prediction~\citep{watter2015embed,song2016linear,kaiser2019model} mitigates representation collapse by forcing the abstract state to be sufficient not just for predicting future \emph{abstract} states, but also future \emph{ground states}.
Unfortunately, in stochastic domains, this comes with the challenging task of density estimation over the ground state space, and as a result, performance is about on-par with end-to-end deep RL~\Citep{vanHasselt2019parametric}.
Moreover, both pixel prediction and the related task of pixel reconstruction~\citep{mattner2012learn,finn2016deep,higgins2017darla,corneil2018efficient,ha2018world,yarats2019improving,hafner2020dream,lee2020slac} are misaligned with the fundamental goal of state abstraction.
These approaches train models to perfectly reproduce the relevant ground state, ergo the abstract state must effectively throw away no information.
By contrast, the objective of state abstraction is to throw away \emph{as much information as possible}, while preserving only what is necessary for decision making. Provided the abstraction is Markov and accurately simulates the ground MDP, we can safely discard the rest of the observation.

\subsection{Inverse Dynamics Models}
As an alternative to (or in addition to) learning a forward model, it is sometimes beneficial to learn an inverse model.
An inverse dynamics model $I(a|x',x)$ predicts the distribution over actions that could have resulted in a transition between a given pair of states. Inverse models have been used for improving generalization from simulation to real-world problems \citep{christiano2016transfer}, enabling effective robot motion planning \citep{agrawal2016learning}, defining intrinsic reward bonuses for exploration \citep{pathak2017curiosity,choi2018contingency}, and decoupling representation learning from rewards \citep{zhang2018decoupling}.
But while inverse models often help with representation learning, we show in Sec. \ref{sec:counterexample} that they are insufficient for ensuring a Markov abstraction.

\subsection{Contrastive Learning}
Since the main barrier to effective next-state prediction is learning an accurate forward model, a compelling alternative is contrastive learning~\citep{gutmann2010noise}, which sidesteps the prediction problem and instead simply aims to decide whether a particular state, or sequence of states, came from one distribution or another.
Contrastive loss objectives typically aim to distinguish either sequential states from non-sequential ones \citep{shelhamer2016loss,anand2019unsupervised, stooke2020decoupling}, real states from predicted ones \citep{oord2018representation}, or determine whether two augmented views came from the same or different observations \citep{laskin2020curl}. Contrastive methods learn representations that in some cases lead to empirically substantial improvements in learning performance, but none has explicitly addressed the question of whether the resulting state abstractions actually preserve the Markov property. We are the first to show that without forward model estimation, pixel prediction/reconstruction, or dependence on reward, the specific combination of inverse model estimation and contrastive learning that we introduce in Section \ref{sec:markov} is sufficient to learn a Markov abstraction.

\subsection{Kinematic Inseparability}
\label{sec:related-work-ki}
One contrastive approach which turns out to be closely related to Markov abstraction is \citeauthor{misra2020kinematic}'s \citeyearpar{misra2020kinematic} HOMER algorithm, and the corresponding notion of \emph{kinematic inseparability} (KI) abstractions.
Two states $x_1'$ and $x_2'$ are defined to be kinematically inseparable if $\Pr(x,a|x'_1) = \Pr(x,a|x'_2)$ and $T(x''|a,x_1')=T(x''|a,x_2')$ (which the authors call ``backwards'' and ``forwards'' KI, respectively).
The idea behind KI abstractions is that unless two states can be distinguished from each other---by either their backward or forward dynamics---they ought to be treated as the same abstract state.
The KI conditions are slightly stronger than the ones we describe in Section \ref{sec:markov}, although when we convert our conditions into a training objective in Section \ref{sec:training}, we additionally satisfy a novel form of the KI conditions, which helps to prevent representation collapse.
While our approach works for both continuous and discrete state spaces, HOMER was only designed for discrete abstract states, and requires specifying---in advance---an upper bound on the \emph{number} of abstract states (which is impossible for continuous state spaces), as well as learning a ``policy cover'' to reach each of those abstract states (which remains impractical even under discretization).

(For a more detailed discussion about KI and Markov abstractions, see Appendix \appref{appendix:KI}{F}.)

\subsection{Other Approaches}
\label{sec:bg-smoothness}
The Markov property is just one of many potentially desirable properties that a representation might have. Not all Markov representations are equally beneficial for learning; otherwise, simply training an RL agent end-to-end on (frame-stacked) image inputs ought to be sufficient, and none of the methods in this section would need to do representation learning at all.

Smoothness is another desirable property and its benefits in RL are well known \citep{pazis2013pac,pirotta2015policy,asadi2018lipschitz}.
Both DeepMDP \citep{gelada2019deepmdp} and DBC \citep{zhang2021learning}, which we compare against, utilize Lipschitz smoothness when learning abstract state representations.
We find in Section \ref{sec:online} that a simple smoothness objective helps our approach in a similar way.
A full investigation of other representation-learning properties (e.g. value preservation \citep{abel2016near}, symbol construction \citep{konidaris2018skills}, suitability for planning \citep{kurutach2018plannable}, information compression \citep{abel2019rlit}) is beyond the scope of this paper.

Since our approach does not require any reward information and is agnostic as to the underlying RL algorithm, it would naturally complement exploration methods designed for sparse reward problems \citep{pathak2017curiosity,burda2018exploration}.
Exploration helps to ensure that the experiences used to learn the abstraction cover as much of the ground MDP's state space as possible.
In algorithms like HOMER (above) and the more recent Proto-RL \citep{yarats2021reinforcement}, the exploration and representation learning objectives are intertwined, whereas our approach is, in principle, compatible with any exploration algorithm. Here we focus solely on the problem of learning Markov state abstractions and view exploration as an exciting direction for future work.




\section{Markov State Abstractions}
\label{sec:markov}
Recall that for a state representation to be Markov (whether ground or abstract), it must be a sufficient statistic for predicting the next state and expected reward, for any action the agent selects.
The state representation of the ground MDP is Markov by definition, but learned state abstractions typically have no such guarantees.
In this section, we introduce conditions that provide the missing guarantees.

Accurate abstract modeling the ground MDP requires replacing the fixed weighting scheme $w(x)$ of Section \ref{sec:background} with a belief distribution, denoted by $B_\phi(x|\{\cdots\})$, that measures the probability of each ground state $x$, conditioned on the entire history of agent experiences. Our objective is to find an abstraction $\phi$ such that any amount of history can be summarized with a single abstract state $z$.

When limited to the most recent abstract state $z$, $B_\phi$ may be policy-dependent and non-stationary:\footnote{This section closely follows \citet{hutter2016extreme}, except here we consider belief distributions over ground states, rather than full histories. An advantage of Hutter's work is that it also considers \emph{abstractions} over histories, though it only provides a rough sketch of how to learn such abstractions. Extending our learning objective to support histories is a natural direction for future work.}
\[ B_{\phi,t}^\pi(x|z) := {\frac{\ind[\phi(x)=z]\ P_t^{\pi}(x)}{\sum_{\tilde{x} \in z} P_t^{\pi}(\tilde{x})}} \taghere \label{eqn:B_phi}, \qquad P_t^{\pi}(x) := \sum_{a\in A} \sum_{\tilde x\in X}T(x | a, \tilde x) \pi_{t-1}(a|\tilde x) P_{t-1}^{\pi}(\tilde x),\]
for $t\ge1$, where $\ind[\cdot]$ denotes the indicator function, $\pi$ is the agent's (possibly non-stationary) behavior policy, and $P_0$ is an arbitrary initial state distribution. Note that $P_t^{\pi}$ and $B_{\phi,t}^\pi$ may still be non-stationary even if $\pi$ is stationary.\footnote{This can happen, for example, when the policy induces either a Markov chain that does not have a stationary distribution, or one whose stationary distribution is different from $P_0$.}

We generalize to a $k$-step belief distribution (for $k \ge 1$) by conditioning \eqref{eqn:B_phi} on additional history:
\begin{align*}
	&B_{\phi,t}^{\pi(k)}\left(x_t | z_t, \{a_{t-i}, z_{t-i}\}_{i=1}^{k}\right) := \\
	&\hspace{35px}{\frac{\ind[\phi(x_t)=z_t]\ \sum_{x_{t-1}\in X}\ T(x_t | a_{t-1}, x_{t-1})\,  B_{\phi,t}^{\pi (k-1)}\left(x_{t-1} \mid z_{t-1},\{a_{t-i},z_{t-i}\}_{i=2}^{k}\right)}
	{\sum_{\tilde{x}_t\in z_t} \sum_{\tilde{x}_{t-1}\in z_{t-1}}\ T(\tilde{x}_t | a_{t-1}, \tilde{x}_{t-1})\, B_{\phi,t}^{\pi (k-1)}\left(\tilde{x}_{t-1} \mid z_{t-1},\{a_{t-i},z_{t-i}\}_{i=2}^{k}\right)}},
\taghere \label{eqn:B_k}
\end{align*}%
where $B_{\phi,t}^{\pi(0)} := B_{\phi,t}^\pi$. Any abstraction induces a belief distribution, but the latter is only independent of history for a \emph{Markov} abstraction. We formalize this concept with the following definition.

\begin{definition}[Markov State Abstraction]
\label{def:abstract-markov}
Given an MDP $M = (X, A, R, T, \gamma)$, initial state distribution $P_0$, and policy class $\Pi_C$, a state abstraction $\phi:X\rightarrow Z$ is \emph{Markov} if and only if for any policy $\pi \in \Pi_C$, $\phi$ induces a belief distribution $B_\phi^\pi$ such that for all $x\in X$, $z\in Z$, $a\in A$, and $k \ge 1$: %
	$ B^{\pi (k)}_{\phi,t}\left(x|z_t,\{a_{t-i},z_{t-i}\}_{i=1}^{k}\right) = B_{\phi,t}^\pi\left(x|z_t\right). $
\end{definition}

In other words, Markov abstractions induce belief distributions that only depend on the most recent abstract state. This property allows an agent to avoid belief distributions entirely, and base its decisions solely on abstract states. Note that Definition \ref{def:abstract-markov} is stricter than Markov state representations (Def. \ref{def:ground-markov}). An abstraction that collapses every ground state to a single abstract state still produces a Markov state representation, but for non-trivial ground MDPs it also induces a history-dependent belief distribution.

Given these definitions, we can define the abstract transitions and rewards for the policy class $\Pi_\phi$ (see Eqn. \eqref{eqn:abstract-policy}) as follows:\footnote{For the more general definitions that support arbitrary policies, see Appendix \appref{appendix:general-policy-defs}{C}.}
\begin{align*}
	T_{\phi,t}^\pi(z'|a,z) &= \sum_{x' \in z'}\ \sum_{x\in z} T(x'|a,x) B_{\phi,t}^\pi(x|z), \taghere \label{eqn:T_phi}\\
	R_{\phi,t}^\pi(z',a,z) &= \sum_{x' \in z'}\ \sum_{x\in z}  {\frac{R(x',a,x)T(x'|a,x)}{T_{\phi,t}^\pi(z'|a,z)}B_{\phi,t}^\pi(x|z)}.\taghere \label{eqn:R_phi}
\end{align*}
Conditioning the belief distribution on additional history yields $k$-step versions compatible with Definition \ref{def:ground-markov}. In the special case where $B_\phi(x|z)$ is stationary and policy-independent (and if rewards are defined over state-action pairs), we recover the fixed weighting function $w(x)$ of \citet{li2006towards}.

\subsection{Sufficient Conditions for a Markov Abstraction}
\label{sec:sufficient-conditions}
The strictly necessary conditions for ensuring an abstraction $\phi$ is Markov over its policy class $\Pi_\phi$ depend on $T$ and $R$, which are typically unknown and hard to estimate due to $X$'s high-dimensionality. However, we can still find sufficient conditions without explicitly knowing $T$ and $R$. To do this, we require that two quantities are equivalent in $M$ and $M_\phi$: the inverse dynamics model, and a density ratio that we define below. The inverse dynamics model $I_t^\pi(a|x',x)$ is defined in terms of the transition function $T(x'|a, x)$ and expected next-state dynamics $P_t^\pi(x'|x)$ via Bayes' theorem: $I_t^\pi(a|x',x) := \frac{T(x'|a, x) \pi_{t}(a|x)}{P_t^\pi(x'|x)}$, where $P_t^\pi(x'|x) = \sum_{\tilde{a} \in A} T(x'|\tilde{a}, x) \pi_t(\tilde{a}|x)$. The same is true of their abstract counterparts, $I^\pi_{\phi,t}(a|z',z)$ and $P_{\phi,t}^\pi(z'|z)$.

\begin{theorem}
\label{thm:markov-conditions}
If $\phi:X\rightarrow Z$ is a state abstraction of MDP $M = (X,A,R,T,\gamma)$ such that for any policy $\pi$ in the policy class $\Pi_\phi$, the following conditions hold for every timestep $t$:
\begin{enumerate}
    \item \textbf{Inverse Model.} The ground and abstract inverse models are equal: $I_{\phi,t}^\pi(a|z',z) = I_t^\pi(a|x',x)$, for all $a \in A$; $z,z' \in Z$; $x,x'\in X$, such that $\phi(x)=z$ and $\phi(x')=z'$. \label{eqn:match-inverse}
    \item \textbf{Density Ratio.} The ground and abstract next-state density ratios are equal, when conditioned on the same abstract state: $\frac{P^\pi_t(z'|z)}{P^\pi_t(z')} = \frac{P^\pi_t(x'|z)}{P^\pi_t(x')}$, for all $z,z'\in Z$; $x'\in X$, such that $\phi(x')=z'$, where $P^\pi_t(x'|z) = \sum_{\tilde x \in X} P^\pi_t(x'|\tilde x)B^\pi_{\phi,t}(\tilde x|z)$, and $P^\pi_t(z') = \sum_{\tilde x' \in z'} P^\pi_t(\tilde x')$.
\end{enumerate}
Then $\phi$ is a Markov state abstraction.
\end{theorem}

\begin{corollary}
\label{corollary:markov-B-implies-MDP}
If $\phi:X\rightarrow Z$ is a Markov state abstraction of MDP $M = (X, A, R, T, \gamma)$ over the policy class $\Pi_\phi$, then the abstract decision process $M_\phi = (Z, A, R^\pi_{\phi,t}, T^\pi_{\phi,t}, \gamma)$ is also Markov.
\end{corollary}

We defer all proofs to Appendix \appref{appendix:proofs}{D}.

Theorem \ref{thm:markov-conditions} describes a pair of conditions under which $\phi$ is a Markov abstraction. Of course, the conditions themselves do not constitute a training objective---we can only use them to confirm an abstraction is Markov. In Section \ref{sec:training}, we adapt these conditions into a practical representation learning objective that is differentiable and suitable for learning $\phi$ using deep neural networks. First, we show why the Inverse Model condition alone is insufficient.

\subsection{An Inverse Model Counterexample}
\label{sec:counterexample}
The example MDP in Figure \ref{fig:inverse-counterexample} additionally demonstrates why the Inverse Model condition alone is insufficient to produce a Markov abstraction. Observe that any valid transition between two ground states uniquely identifies the selected action. The same is true for abstract states since the only way to reach $z_B$ is via action $a_1$, and the only way to leave is action $a_0$. Therefore, the abstraction satisfies the Inverse Model condition for any policy. However, as noted in Section \ref{sec:bg-state-abstraction}, conditioning on additional history changes the abstract transition probabilities, and thus the Inverse Model condition is not sufficient for an abstraction to be Markov. In fact, we show in Appendix \appref{appendix:inverse-implies-ratio}{D.2} that, given the Inverse Model condition, the Density Ratio condition is actually \emph{necessary} for a Markov abstraction.

\section{Training a Markov Abstraction}
\label{sec:training}
We now present a set of training objectives for approximately satisfying the conditions of Theorem \ref{thm:markov-conditions}. Since the theorem applies for the policy class $\Pi_\phi$ induced by the abstraction, we restrict the policy by defining $\pi$ as a mapping from $Z\rightarrow \Pr(A)$, rather than from $X \rightarrow \Pr(A)$. In cases where $\pi$ is defined implicitly via the value function, we ensure that the latter is defined over abstract states.

\paragraph{Inverse Models.}
To ensure the ground and abstract inverse models are equal, we consider a batch of $N$ experiences $(x_i, a_i, x_i')$, encode ground states with $\phi$, and jointly train a model $f(a|\phi(x_i'),\phi(x_i);\theta_f)$ to predict a distribution over actions, with $a_i$ as the label. This can be achieved by minimizing a cross-entropy loss, for either discrete or continuous action spaces:
\[\mathcal{L}_{Inv} := -\frac{1}{N}\sum_{i=1}^N \log f(a=a_i|\phi(x_i'), \phi(x_i);\theta_f).\notag \]
Note that because the policy class is restricted to $\Pi_\phi$, if the policy is stationary and deterministic, then $I_{\phi,t}^\pi(a|z',z) = \pi_{\phi}(a|z) = \pi(a|x) = I_t^\pi(a|x',x)$ and the Inverse Model condition is satisfied trivially. Thus we expect $\mathcal{L}_{Inv}$ to be most useful for representation learning when the policy has high entropy or is changing rapidly, such as during early training.

\paragraph{Density Ratios.} The second condition, namely that $\frac{P_{\phi,t}^\pi(z'|z)}{P_{\phi,t}^\pi(z')} = \frac{P_t^\pi(x'|z)}{P_t^\pi(x')}$, means we can distinguish conditional samples from marginal samples equally well for abstract states or ground states. This objective naturally lends itself to a type of contrastive loss. We generate a batch of $N$ sequential state pairs $(x_i, x_i')$ as samples of $\Pr(x'|x)$, and a batch of $N$ non-sequential state pairs $(x_i, \tilde x_i')$ as samples of $\Pr(x')$, where the latter pairs can be obtained, for example, by shuffling the $x_i'$ states in the first batch. We assign positive labels ($y_i = 1$) to sequential pairs and negative labels to non-sequential pairs. This setup, following the derivation of \citet{tiao2017simple}, allows us to write density ratios in terms of class-posterior probabilities: $\delta(x') := \frac{\Pr(x'|x)}{\Pr(x')} = \frac{p(y=1|x,x')}{1-p(y=1|x,x')}$ and $\delta_\phi(z') := \frac{\Pr(z'|z)}{\Pr(z')} = \frac{q(y=1|z, z')}{1-q(y=1|z, z')}$, where $p$ and $q$ are just names for specific probability distributions.\footnote{For completeness, we reproduce the derivation in Appendix \appref{appendix:ratio-derivation}{E}.}
We jointly train an abstraction $\phi$ and a classifier $g(y|\phi(x'),\phi(x);\theta_g)$, minimizing the cross-entropy between predictions and labels $y_i$:
\[\mathcal{L}_{Ratio} := -\frac{1}{2N} \sum_{i=1}^{2N} \log g(y=y_i|\phi(x_i'), \phi(x_i); \theta_g). \notag\]
In doing so, we ensure $g$ approaches $p$ and $q$ simultaneously, which drives $\delta_\phi(z') \rightarrow \delta(x')$.

Note that this is stronger than the original condition, which only required the ratios to be equal in expectation.
This stronger objective, when combined with the inverse loss, actually encodes a novel form of the kinematic inseparability conditions from Section \ref{sec:related-work-ki}, which further helps to avoid representation collapse. (See Appendix \appref{appendix:KI}{F} for more details.)

\paragraph{Smoothness.}
To improve robustness and encourage our method to learn smooth representations like those discussed in Section \ref{sec:bg-smoothness}, we optionally add an additional term to our loss function:
\[\mathcal{L}_{Smooth} := (\text{ReLU}(\|\phi(x') - \phi(x)\|_2 - d_0))^2.\]
This term penalizes consecutive abstract states for being more than some predefined distance $d_0$ away from each other.
Appendix \appref{appendix:smoothness}{L} describes an additional experiment and provides further justification for why representation smoothness is an important consideration that complements the Markov property.

\paragraph{Markov Abstraction Objective.} We generate a batch of experiences using a mixture of abstract policies $\pi_i \in \Pi_C \subseteq \Pi_\phi$ (for example, with a uniform random policy), then train $\phi$ end-to-end while minimizing a weighted combination of the inverse, ratio, and smoothness losses:
\[\mathcal{L}_{Markov} := \alpha \mathcal{L}_{Inv} + \beta \mathcal{L}_{Ratio}  + \eta \mathcal{L}_{Smooth},\]
where $\alpha$, $\beta$, and $\eta$ are coefficients that compensate for the relative difficulty of minimizing each individual objective for the domain in question.

The Markov objective avoids the problem of representation collapse without requiring reward information or ground state prediction. A trivial abstraction like $\phi(x)\mapsto 0$ would not minimize $\mathcal{L}_{Markov}$, because it contains no useful information for predicting actions or distinguishing authentic transitions from manufactured ones.

\section{Offline Abstraction Learning for Visual Gridworlds}
\label{sec:offline}
First, we evaluate our approach for learning an abstraction offline for a visual gridworld domain (Fig. \ref{fig:main-objective}, left).
Each discrete $(x,y)$ position in the $6\times 6$ gridworld is mapped to a noisy image (see Appendix \appref{appendix:gridworld}{G}).
We emphasize that the agent only sees these images; it does not have access to the ground-truth $(x,y)$ position.
The agent gathers a batch of experiences in a version of the gridworld with \emph{no rewards or terminal states}, using a uniform random exploration policy over the four directional actions.

\begin{figure}[b]
\begin{subfigure}{.6\textwidth}
    \centering
    \includegraphics[width=.99\columnwidth]{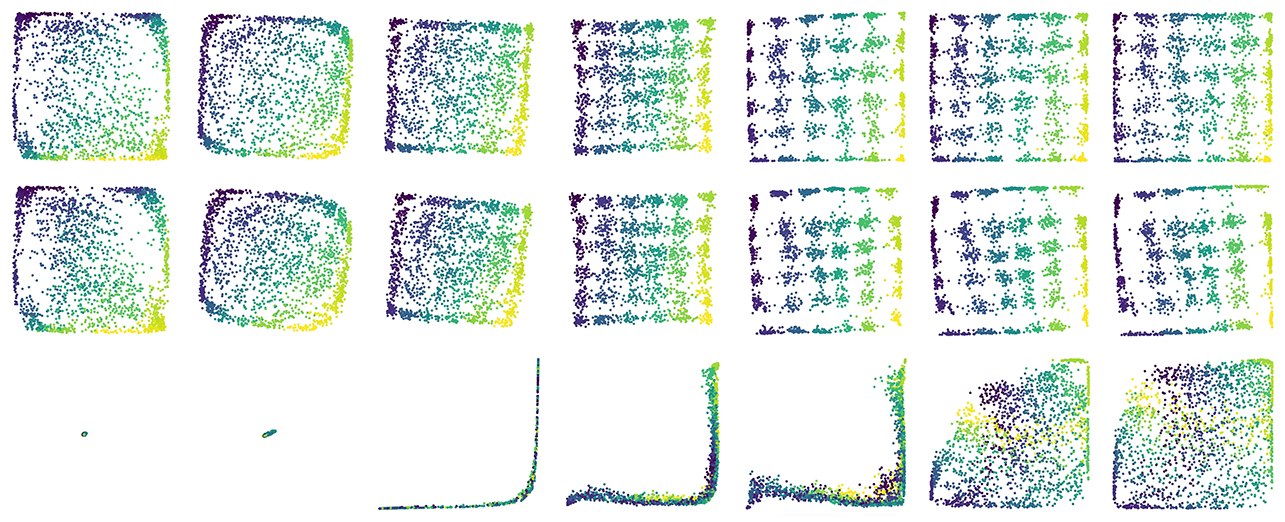}
    \vspace{11px}
    \caption{}
    \label{fig:rep-vis}
\end{subfigure}
\hspace{13px}
\begin{subfigure}{.32\textwidth}
    \includegraphics[width=.99\columnwidth]{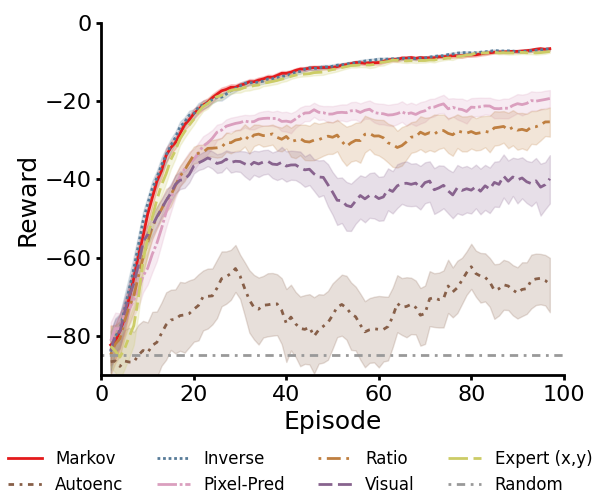}
    \caption{}
    \label{fig:rep-eval}
\end{subfigure}
\caption{(a) Visualization of learning progress at selected times (left to right) of a 2-D state abstraction for the $6 \times 6$ visual gridworld domain: (top row) $\mathcal{L}_{Markov}$; (middle row) $\mathcal{L}_{Inv}$ only; (bottom row) $\mathcal{L}_{Ratio}$ only. Color denotes ground-truth $(x,y)$ position, which is not shown to the agent. (b) Mean episode reward for the visual gridworld navigation task. Markov abstractions significantly outperform end-to-end training with visual inputs, and match the performance of the expert $(x,y)$ position features. (300 seeds; 5-point moving average; shaded regions denote 95\% confidence intervals.)}
\end{figure}

These experiences are then used offline to train an abstraction function $\phi_{Markov}$, by minimizing $\mathcal{L}_{Markov}$ (with $\alpha = \beta = 1$, $\eta=0$).
We visualize the learned 2-D abstract state space in Figure \ref{fig:rep-vis} (top row) and compare against ablations that train with only $\mathcal{L}_{Inv}$ or $\mathcal{L}_{Ratio}$, as well as against two baselines that we train via pixel prediction and reconstruction, respectively (see Appendix \appref{appendix:more-visualizations}{I} for more visualizations).
We observe that $\phi_{Markov}$ and $\phi_{Inv}$ cluster the noisy observations and recover the $6\times 6$ grid structure, whereas the others do not generally have an obvious interpretation.
We also observed that $\phi_{Ratio}$ and $\phi_{Autoenc}$ frequently failed to converge.

Next we froze these abstraction functions and used them to map images to abstract states while training DQN~\citep{mnih2015human} on the resulting features. We measured the learning performance of each pre-trained abstraction, as well as that of end-to-end DQN with no pretraining. We plot learning curves in Figure \ref{fig:rep-eval}. For reference, we also include learning curves for a uniform random policy and DQN trained on ground-truth $(x,y)$ position with no abstraction.

Markov abstractions match the performance of ground-truth position, and beat every other learned representation except $\phi_{Inv}$. Note that while $\phi_{Markov}$ and $\phi_{Inv}$ perform similarly in this domain, there is no reason to expect $L_{Inv}$ to work on its own for other domains, since it lacks the theoretical motivation of our combined Markov loss.
When the combined loss is minimized, the Markov conditions are satisfied. But \emph{even if} the inverse loss goes to zero on its own, the counterexample in Section \ref{sec:counterexample} demonstrates that this is insufficient to learn a Markov abstraction.

\section{Online Abstraction Learning for Continuous Control}
\label{sec:online}
Next, we evaluate our approach in an online setting with a collection of image-based, continuous control tasks from the DeepMind Control Suite \citep{tassa2020dmcontrol}.
Our training objective is agnostic about the underlying RL algorithm, so we use as our baseline the state-of-the-art technique that combines Soft Actor-Critic (SAC) \citep{haarnoja2018soft} with random data augmentation (RAD) \citep{laskin2020reinforcement}.
We initialize a replay buffer with experiences from a uniform random policy, as is typical, but before training with RL, we use those same experiences \emph{with reward information removed} to pretrain a Markov abstraction. We then continue training with the Markov objective alongside traditional RL. (See Appendix \appref{appendix:deepmind-control}{H} for implementation details).

\begin{figure*}[!h]
    \centering
    \includegraphics[width=.92\linewidth]{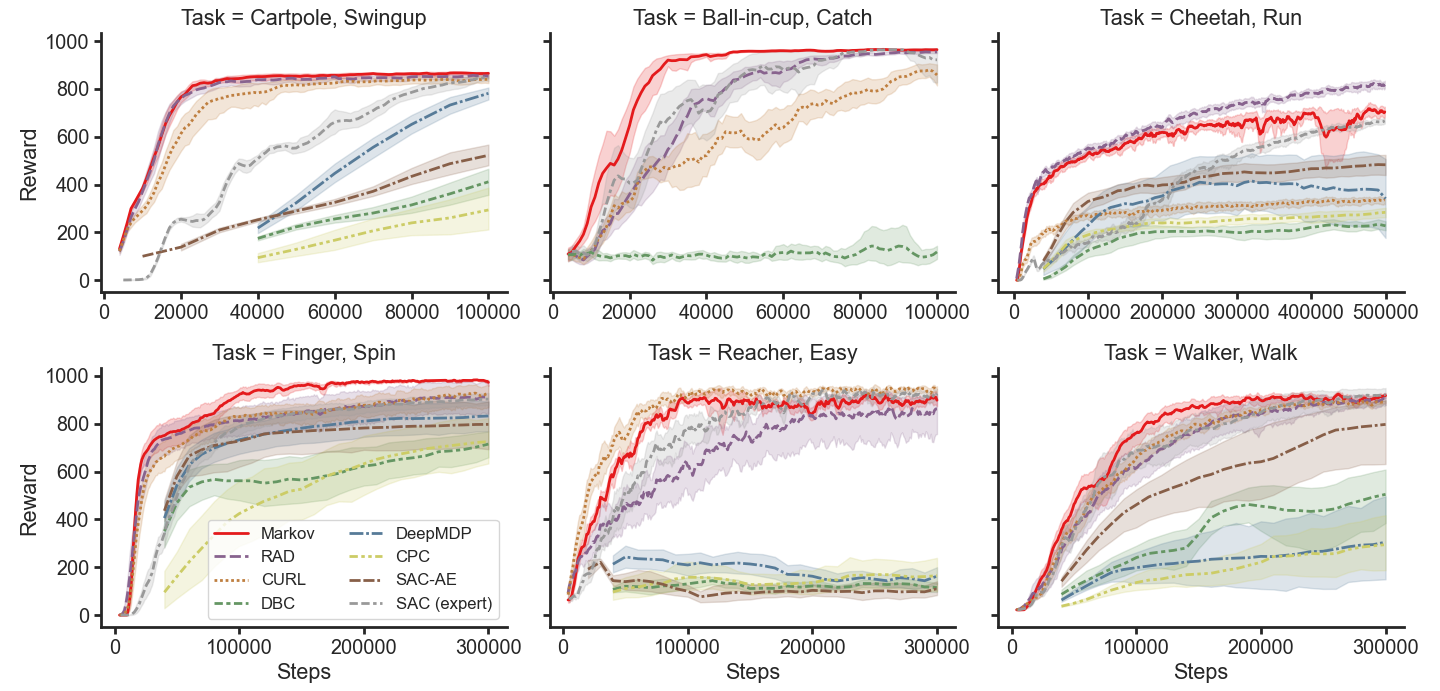}
    \caption{Mean episode reward vs. environment steps for DeepMind Control. Adding our Markov objective leads to improved learning performance. (10 seeds; 5-point moving average; shaded regions denote 90\% confidence intervals; learning curve data is available at the linked code repository.)}
    \label{fig:dmcontrol}
\end{figure*}

In Figure \ref{fig:dmcontrol}, we compare against unmodified RAD, as well as contrastive methods CURL \citep{laskin2020curl} and CPC \citep{oord2018representation}, bisimulation methods DeepMDP \citep{gelada2019deepmdp} and DBC \citep{zhang2021learning}, and pixel-reconstruction method SAC-AE \citep{yarats2019improving}. As a reference, we also include non-visual SAC with expert features.
All methods use the same number of environment steps (the experiences used for pretraining are not additional experiences).

Relative to RAD, our method learns faster on four domains and slower on one, typically achieving the same final performance (better in one, worse in one).
It performs even more favorably relative to the other baselines, of which CURL is most similar to our method, since it combines contrastive learning with data augmentation similar to that of RAD.\footnote{We ran another experiment with no data augmentation, using a different state-of-the-art continuous control algorithm, RBF-DQN \citep{asadi2021deep}, and found similar results there as well (see Appendix \appref{appendix:rbfdqn}{K} for details).}
Our approach even represents a marginal improvement over a hypothetical ``best of'' oracle that always chooses the best performing baseline.
These experiments show that even in an online setting, where the agent can leverage reward information and a Markov ground state when building its abstract state representation, explicitly encouraging Markov abstractions improves learning performance over state-of-the-art image-based RL.

\section{Conclusion}
We have developed a principled approach to learning abstract state representations that provably results in Markov abstract states, and which does not require estimating transition dynamics nor ground-state prediction.
We defined what it means for a state abstraction to be Markov while ensuring that the abstract MDP accurately reflects the dynamics of the ground MDP, and introduced sufficient conditions for achieving such an abstraction.
We adapted these conditions into a practical training objective that combines inverse model estimation and temporal contrastive learning.
Our approach learns abstract state representations, with and without reward, that capture the structure of the underlying domain and substantially improve learning performance over existing approaches.

\ifthenelse{\equal{\includeAppendix}{yes}}{
\begin{ack}
Thank you to Ben Abbatematteo, David Abel, Barrett Ames, S\'eb Arnold, Kavosh Asadi, Akhil Bagaria, Jake Beck, Jules Becker, Alexander Ivanov, Steve James, Michael Littman, Sam Lobel, and our other colleagues at Brown for thoughtful advice and countless helpful discussions, as well as Amy Zhang for generously sharing baseline learning curve data, and the anonymous ICML'21 and NeurIPS'21 reviewers for their time, consideration, and valuable feedback that substantially improved the paper.
This research was supported by the ONR under the PERISCOPE MURI Contract {N00014-17-1-2699}, by the DARPA Lifelong Learning Machines program under grant {FA8750-18-2-0117}, and by NSF grants 1955361, 1717569, and CAREER award 1844960.
\end{ack}
}{}

\section*{Errata}
\begin{enumerate}[leftmargin=20px]
    \item An earlier version of this paper contained a typo in the definition of $P^\pi_t(z')$. The correct definition is $P^\pi_t(z') = \sum_{\tilde x' \in z'} P^\pi_t(\tilde x')$, but originally it read $P^\pi_t(z') = \sum_{\tilde x' \in X} P^\pi_t(\tilde x')B^\pi_{\phi,t}(\tilde x'|z')$.
\end{enumerate}

\DeclareRobustCommand{\VAN}[3]{#3} 

\bibliographystyle{icml2021}
{\small
\bibliography{refs}
}

\appendix
\onecolumn
\etocsettocdepth{2}
\part{Appendix} 
\begingroup
\parindent=0em
\etocsettocstyle{}{}
\localtableofcontents 
\endgroup

\newpage

\section{Broader Impact Statement}
\label{appendix:broader-impact}
Our approach enables agents to automatically construct useful, Markov state representations for reinforcement learning from rich observations. Since most reinforcement learning algorithms implicitly assume Markov abstract state representations, and since agents may struggle to learn when that assumption is violated, this work has the potential to benefit a large number of algorithms.

Our training objective is designed for neural networks, which are not guaranteed to converge to a global optimum when trained with stochastic gradient descent.
Typically, such training objectives will only approximately satisfy the theoretical conditions they encode.
However, this is not a drawback of our method---it applies to any representation learning technique that uses neural networks.
Moreover, as neural network optimization techniques improve, our method will converge to a Markov abstraction, whereas other approaches may not.
In the meantime, systems in safety-critical domains should ensure that they can cope with non-Markov abstractions without undergoing catastrophic failures.

We have shown experimentally that our method is effective in a variety of domains; however, other problem domains may require additional hyperparameter tuning, which can be expensive. Nevertheless, one benefit of our method is that Markov abstractions can be learned offline, without access to reward information. This means our algorithm could be used, in advance, to learn an abstraction for some problem domain, and then subsequent tasks in that environment (perhaps with different reward functions) could avoid the problem of perceptual abstraction as well as any associated training costs.

\newpage
\section{Glossary of Symbols}
\label{appendix:glossary}
Here we provide a glossary of the most commonly-used symbols appearing in the rest of the paper.

\begin{table}[h!]
    \centering
    \small
    \begin{tabular}{ll}
    \toprule
    \textbf{Symbol} & \textbf{Description} \\
    \midrule
    $M = (X,A,R,T,\gamma)$ & Ground MDP\\
    $X$ & Observed (ground) state space \\
    $A$ & Action space\\
    $R(x',a,x)$ & Reward function\\
    $T(x'|a,x)$ & Transition model\\
    $\gamma$ & Discount factor\\
    $R^{(k)}(x',a,x,\{\cdots\}_{i=1}^k)$ & Reward function conditioned on $k$ steps of additional history\\
    $T^{(k)}(x'|a,x,\{\cdots\}_{i=1}^k)$ & Transition model conditioned on $k$ steps of additional history\\
    \midrule
    $S$ & Unobserved (latent) state space, assumed for Block MDPs\\
    $\sigma:S\rightarrow \Pr(X)$ & Sensor function for producing observed states\\
    $\sigma^{-1}:X \rightarrow S$ & Hypothetical ``inverse-sensor'' function, assumed for Block MDPs\\
    \midrule
    $\phi:X\rightarrow Z$ & Abstraction function\\
    $w(x)$ & Fixed ground-state weighting function for constructing an abstract\\
           & MDP \citep{li2006towards}\\
    $B^\pi_{\phi,t}(x|z_t)$ & Belief distribution for assigning policy- and time-dependent\\
           & weights to ground states\\
    $B^{\pi(k)}_{\phi,t}(x|z_t,\{a_{t-i},z_{t-i}\}_{i=1}^{k})$ & Belief distribution conditioned on $k$ steps of history\\
    \midrule
    $M_\phi = (Z, A, R^\pi_{\phi,t}, T^\pi_{\phi,t}, \gamma)$ & Abstract decision process (possibly non-Markov)\\
    $Z$ & Abstract state space\\
    $R^\pi_{\phi,t}(z',a,z)$ & Abstract reward function\\
    $T^\pi_{\phi,t}(z'|a,z)$ & Abstract transition model\\
    $R^{\pi(k)}_{\phi,t}(z',a,z,\{\cdots\}_{i=1}^{k})$ & Abstract reward function conditioned on $k$ steps of additional history\\
    $T^{\pi(k)}_{\phi,t}(z'|a,z,\{\cdots\}_{i=1}^{k})$ & Abstract transition model conditioned on $k$ steps of additional history\\
    \midrule
    $\pi(a|x)$ & A policy\\
    $\pi^*$ &An optimal policy\\
    $\Pi_C$ & An arbitrary policy class\\
    $\Pi_\phi$ & The class of abstract policies induced by abstraction $\phi$\\
    $V^\pi: X \rightarrow \mathbb{R}$ & The value function induced by $\pi$\\
    \midrule
    $P^\pi_{t}(x)$ & Ground-state visitation distribution\\
    $P_t^\pi(x'|x)$ & Expected next-state dynamics model\\
    $I^\pi_t(a|x',x)$ & Inverse dynamics model\\
    $P^\pi_{\phi,t}(z)$ & Abstract-state visitation distribution\\
    $P^\pi_{\phi,t}(z'|z)$ & Abstract expected next-state dynamics model\\
    $I^\pi_{\phi,t}(a|z',z)$ & Abstract inverse dynamics model\\
    \midrule
    $P_t^\pi(x,a|x')$ & Backwards dynamics model, used for kinematic inseparability\\
    \bottomrule
    \end{tabular}
    \caption{Glossary of symbols}
    \label{tab:glossary}
\end{table}

\newpage
\section{General-Policy Definitions}
\label{appendix:general-policy-defs}

\subsection{General-Policy Definitions}
The definitions of $T^\pi_{\phi,t}$ and $R^\pi_{\phi,t}$ in Section \ref{sec:markov} only apply when the policy $\pi$ is a member of $\Pi_\phi$. Here we derive more general definitions that are valid for arbitrary policies, not just those in $\Pi_\phi$. For the special case where $\pi \in \Pi_\phi$, these definitions are equivalent to equations \eqref{eqn:T_phi} and \eqref{eqn:R_phi}.

\subsubsection*{Abstract transition probabilities.}

\begin{align*}
		&\hphantom{:=}\ \Pr(z'| a, z) \\
		&\hphantom{:}=\ \sum_{x'\in X}\Pr(z'|x', a, z)\Pr(x'|a,z)\\
		&\hphantom{:}=\ \sum_{x'\in X:\phi(x')=z'}\Pr(x'|a, z)\\
		&\hphantom{:}=\ \sum_{x'\in X:\phi(x')=z'}\ \sum_{x\in X} \Pr(x'|a, x, z)\Pr(x|a, z)\\
		&\hphantom{:}=\ \sum_{x'\in X:\phi(x')=z'}\ \sum_{x\in X} \Pr(x'|a, x, z) \frac{\Pr(a|x, z)\Pr(x|z)}{\sum_{\tilde{x} \in X} \Pr(a|\tilde{x}, z) \Pr(\tilde{x}_t|z_t)}\\
	T_{\phi,t}^\pi(z'| a, z) &:=\ \sum_{x'\in X:\phi(x')=z'}\ \sum_{x\in X:\phi(x)=z} T(x'|a, x) \frac{\pi_t(a|x) B_{\phi,t}^\pi(x|z)}{\sum_{\tilde{x} \in X} \pi_t(a|\tilde{x}) B_{\phi,t}^\pi(\tilde{x}|z)}
\end{align*}

\subsubsection*{Abstract rewards.}

\begin{align*}
	&\hphantom{:=} \sum_{r\in R} r \Pr(r|z', a, z)\\
	&\hphantom{:}=\ \sum_{x'\in X}\ \sum_{x\in X}\ \sum_{r\in R} r \Pr(r,x',x|z', a, z)\\
	&\hphantom{:}=\ \sum_{x'\in X}\ \sum_{x\in X}\ \sum_{r\in R} r \Pr(r|x', z', a, x, z) \Pr(x', x|z', a, z)\\
	&\hphantom{:}= \ \sum_{x'\in X}\sum_{x\in X} R(x', a, x)\frac{\Pr(z'| x', a, x, z) \Pr(x',x|a,z)}{\Pr(z'|a, z)}\\
	&\hphantom{:}=\ \sum_{x'\in X:\phi(x')=z'}\ \sum_{x \in X} R(x', a, x) \frac{\Pr(x'| a, x, z) \Pr(x|a, z)}{\Pr(z'|a, z)}\\
	&\hphantom{:}=\ \sum_{x'\in X:\phi(x')=z'}\ \sum_{x\in X} R(x', a, x)\frac{\Pr(x'|a, x)}{\Pr(z'|a,z)}\frac{\Pr(a|x)\Pr(x|z)}{\sum_{\tilde{x}\in X} \Pr(a|\tilde{x}) \Pr(\tilde{x}|z)}\\
	R_{\phi,t}^\pi(z',a,z) &:=\ \sum_{x'\in X:\phi(x')=z'}\ \sum_{x\in X:\phi(x)=z} R(x', a, x)\frac{T(x'|a, x)\, \pi_t(a|x)B_{\phi,t}^\pi(x|z)}{T_{\phi,t}^\pi(z'|a,z)\sum_{\tilde{x}\in X}\pi_t(a|\tilde{x})B_{\phi,t}^\pi(\tilde{x}|z)}
\end{align*}

\newpage
\section{Proofs}
\label{appendix:proofs}

Here we provide proofs of Theorem \ref{thm:markov-conditions} and its corollary, which state that the Inverse Model and Density Ratio conditions are sufficient for $\phi$ and $M_\phi$ to be Markov. Then, to complement the counterexample from Section \ref{sec:counterexample}, we also present and prove a second theorem which states that, given the Inverse Model condition, the Density Ratio condition is in fact \emph{necessary} for a Markov abstraction.

\subsection{Main Theorem}

The proof of Theorem \ref{thm:markov-conditions} makes use of two lemmas: Lemma \ref{lemma:B_step_to_TR_step}, that equal $k$-step and ($k-1$)-step belief distributions imply equal $k$-step and ($k-1$)-step transition models and reward functions, and Lemma \ref{lemma:B_step_to_B_step}, that equal $k$-step and ($k-1$)-step belief distributions imply equal ($k+1$)-step and $k$-step belief distributions. Since the lemmas apply for any arbitrary policy, we use the general-policy definitions from Appendix \ref{appendix:general-policy-defs}.

\begin{lemma}
\label{lemma:B_step_to_TR_step}
	Given an MDP $M$, abstraction $\phi$, policy $\pi$, initial state distribution $P_0$, and any $k\ge 1$, if $B_{\phi,t}^{\pi(k)}(x_t|z_t,\{a_{t-i},z_{t-i}\}_{i=1}^k) = B_{\phi,t}^{\pi (k-1)}(x_t|z_t,\{a_{t-i},z_{t-i}\}_{i=1}^{k-1})$, then for all $a_t \in A$ and $z_{t+1}\in Z$:
	\begin{align*}
		 T_{\phi,t}^{\pi (k)}(z_{t+1} | \{a_{t-i}, z_{t-i}\}_{i=0}^{k}) &= T_{\phi,t}^{\pi(k-1)}(z_{t+1} | \{a_{t-i}, z_{t-i}\}_{i=0}^{k-1})\\
		\mathlarger{\cap}\quad  R_{\phi,t}^{\pi (k)}(z_{t+1},\{a_{t-i}, z_{t-i}\}_{i=0}^{k}) &= R_{\phi,t}^{\pi(k-1)}(z_{t+1},\{a_{t-i}, z_{t-i}\}_{i=0}^{k-1}).
	\end{align*}
\end{lemma}
In the proof below, we start with $B_{\phi,t}^{\pi(k)} = B_{\phi,t}^{\pi (k-1)}$, and repeatedly multiply or divide both sides by the same quantity, or take the same summations of both sides, to obtain $T_{\phi,t}^{\pi (k)} = T_{\phi,t}^{\pi(k-1)}$, then apply the same process again, making use of the fact that $T_{\phi,t}^{\pi (k)} = T_{\phi,t}^{\pi(k-1)}$, to obtain $R_{\phi,t}^{\pi(k)} = R_{\phi,t}^{\pi(k-1)}$.

\begin{Proof}\hspace{10px}

	\begin{align*}
		&\hphantom{\Rightarrow} & B_{\phi,t}^{\pi(k)}\left(x_t\middle|z_t,\{a_{t-i},z_{t-i}\}_{i=1}^k\right) &= B_{\phi,t}^{\pi (k-1)}\left(x_t\middle|z_t,\{a_{t-i},z_{t-i}\}_{i=1}^{k-1}\right).\\
		\shortintertext{Let $a_t \in A$ be any action.}
		&\Rightarrow &\pi_t(a_t|x_t) B_{\phi,t}^{\pi(k)}\left(x_t\middle|z_t,\{a_{t-i},z_{t-i}\}_{i=1}^k\right) &= \pi_t(a_t|x_t) B_{\phi,t}^{\pi (k-1)}\left(x_t\middle|z_t,\{a_{t-i},z_{t-i}\}_{i=1}^{k-1}\right)\\
		&\Rightarrow &\mathsmaller{\frac{\pi_t(a_t|x_t) B_{\phi,t}^{\pi(k)}\left(x_t\middle|z_t,\{a_{t-i},z_{t-i}\}_{i=1}^k\right)}{\sum_{\tilde{x}_t\in X} \pi_t(a_t|\tilde{x}_t) B_{\phi,t}^{\pi(k)}\left(\tilde{x}_t\middle|z_t,\{a_{t-i},z_{t-i}\}_{i=1}^k\right)}}
			&= \mathsmaller{\frac{\pi_t(a_t|x_t) B_{\phi,t}^{\pi (k-1)}\left(x_t\middle|z_t,\{a_{t-i},z_{t-i}\}_{i=1}^{k-1}\right)}{\sum_{\tilde{x}_t\in X} \pi_t(a_t|\tilde{x}_t) B_{\phi,t}^{\pi (k-1)}\left(\tilde{x}_t\middle|z_t,\{a_{t-i},z_{t-i}\}_{i=1}^{k-1}\right)}}. \taghere \label{eqn:B_step_to_TR_step_pt1}
	\end{align*}
	Let
	\begin{equation}
		C_{\phi,t}^{\pi (k)} := \mathsmaller{\frac{\pi_t(a_t|x_t) B_{\phi,t}^{\pi(k)}\left(x_t \mid z_t,\{a_{t-i},z_{t-i}\}_{i=1}^k\right)}{\sum_{\tilde{x}_t\in X} \pi_t(a_t|\tilde{x}_t) B_{\phi,t}^{\pi(k)}\left(\tilde{x}_t \mid z_t,\{a_{t-i},z_{t-i}\}_{i=1}^k\right)}} \label{eqn:C_k}
	\end{equation}

	Combining \eqref{eqn:B_step_to_TR_step_pt1} and \eqref{eqn:C_k}, we obtain:
	\begin{align*}
		&\hphantom{\Rightarrow} & C_{\phi,t}^{\pi (k)} &= C_{\phi,t}^{\pi (k-1)}\taghere \label{eqn:C_step}\\
		&\Rightarrow &\sum_{x_t \in z_t}\ \sum_{x_{t+1}\in z_{t+1}}\, T\left(x_{t+1} \mid a_t,x_t\right)C_{\phi,t}^{\pi (k)}
		&= \sum_{x_t \in z_t}\ \sum_{x_{t+1}\in z_{t+1}}\, T\left(x_{t+1} \mid a_t,x_t\right)C_{\phi,t}^{\pi (k-1)}\\
		&\Leftrightarrow & T_{\phi,t}^{\pi (k)}\left(z_{t+1} \mid \{a_{t-i}, z_{t-i}\}_{i=0}^{k}\right) &= T_{\phi,t}^{\pi(k-1)}\left(z_{t+1} \mid \{a_{t-i}, z_{t-i}\}_{i=0}^{k-1}\right). \taghere \label{eqn:T_step}
	\end{align*}

	Additionally, we can combine \eqref{eqn:C_step} and \eqref{eqn:T_step} and apply the same approach for rewards:
	\begin{align*}
		&\hphantom{\Rightarrow} & C_{\phi,t}^{\pi (k)} &= C_{\phi,t}^{\pi (k-1)}\\
		&\Rightarrow & \mathsmaller{\frac{C_{\phi,t}^{\pi (k)}}{ T_{\phi,t}^{\pi (k)}(z_{t+1} | \{a_{t-i}, z_{t-i}\}_{i=0}^{k})}} &= \mathsmaller{\frac{C_{\phi,t}^{\pi (k-1)}}{T_{\phi,t}^{\pi(k-1)}(z_{t+1} | \{a_{t-i}, z_{t-i}\}_{i=0}^{k-1})}}\\
		&\Rightarrow & \mathsmaller{\frac{T(x_{t+1}|a_t,z_t)C_{\phi,t}^{\pi (k)}}{ T_{\phi,t}^{\pi (k)}(z_{t+1} | \{a_{t-i}, z_{t-i}\}_{i=0}^{k})}} &= \mathsmaller{\frac{T(x_{t+1}|a_t,z_t)C_{\phi,t}^{\pi (k-1)}}{T_{\phi,t}^{\pi(k-1)}(z_{t+1} | \{a_{t-i}, z_{t-i}\}_{i=0}^{k-1})}}\\
		&\Rightarrow & R_{\phi,t}^{\pi(k)}\left(z_{t+1},\{a_{t-i},z_{t-i}\}_{i=0}^k\right) &= R_{\phi,t}^{\pi(k-1)}\left(z_{t+1},\{a_{t-i},z_{t-i}\}_{i=0}^{k-1}\right). \taghere \label{eqn:R_step}
	\end{align*}
\end{Proof}

%

\begin{lemma}
\label{lemma:B_step_to_B_step}
Given an MDP $M$, abstraction $\phi$, policy $\pi$, and initial state distribution $P_0$, if for all $t \ge k$, $z_t\in Z$, and $x_t\in X$ such that $\phi(x_t)=z_t$, it holds that $B_{\phi,t}^{\pi(k)}(x_t|z_t,\{a_{t-i},z_{t-i}\}_{i=1}^k) = B_{\phi,t}^{\pi (k-1)}(x_t|z_t,\{a_{t-i},z_{t-i}\}_{i=1}^{k-1})$, then for all $z_{t+1}\in Z$, $x_{t+1} \in X: \phi(x_{t+1})=z_{t+1}$,
	\[\begin{aligned}
	    &B^{\pi(k+1)}_{\phi,t}\left(x_{t+1}\middle|z_{t+1},\{a_{t-i},z_{t-i}\}_{i=0}^{k}\right) = B_{\phi,t}^{\pi(k)}\left(x_{t+1}\middle|z_{t+1},\{a_{t-i},z_{t-i}\}_{i=0}^{k-1}\right).
	\end{aligned}\]
\end{lemma}

To prove this lemma, we invoke Lemma \ref{lemma:B_step_to_TR_step} to obtain $ T_{\phi,t}^{\pi (k)} = T_{\phi,t}^{\pi(k-1)}$, and then follow the same approach as before, performing operations to both sides until we achieve the desired result.

\begin{Proof}\hspace{10px}

    Let $ T_{\phi,t}^{\pi (k)}$ be defined via \eqref{eqn:T_phi} and \eqref{eqn:B_k}. Applying Lemma \ref{lemma:B_step_to_TR_step} to the premise gives:
    \[ T_{\phi,t}^{\pi (k)}(z_{t+1} | \{a_{t-i}, z_{t-i}\}_{i=0}^{k}) = T_{\phi,t}^{\pi(k-1)}(z_{t+1} | \{a_{t-i}, z_{t-i}\}_{i=0}^{k-1}). \]
    Returning to the premise, we have:
	\begin{align*}
		&\hphantom{\Rightarrow} & B_{\phi,t}^{\pi(k)}\left(x_t\middle|z_t,\{a_{t-i},z_{t-i}\}_{i=1}^k\right) &= B_{\phi,t}^{\pi (k-1)}\left(x_t\middle|z_t,\{a_{t-i},z_{t-i}\}_{i=1}^{k-1}\right)\\
		&\Rightarrow & \frac{B_{\phi,t}^{\pi(k)}\left(x_t\middle|z_t,\{a_{t-i},z_{t-i}\}_{i=1}^k\right)} { T_{\phi,t}^{\pi (k)}\left(z_{t+1} \middle| \{a_{t-i}, z_{t-i}\}_{i=0}^{k}\right)} &= \frac{B_{\phi,t}^{\pi (k-1)}\left(x_t\middle|z_t,\{a_{t-i},z_{t-i}\}_{i=1}^{k-1}\right)}{T_{\phi,t}^{\pi(k-1)}\left(z_{t+1} \middle| \{a_{t-i}, z_{t-i}\}_{i=0}^{k-1}\right)}\\
		&\Rightarrow & \sum_{\mathclap{\substack{x_t\in X:\\ \phi(x_t)=z_t}}} \mathsmaller{\frac{T(x_{t+1}|a_t,x_t) B_{\phi,t}^{\pi(k)}\left(x_t\middle|z_t,\{a_{t-i},z_{t-i}\}_{i=1}^k\right)}{ T_{\phi,t}^{\pi (k)}\left(z_{t+1} \middle| \{a_{t-i}, z_{t-i}\}_{i=0}^{k}\right)}} &= \sum_{\mathclap{\substack{x_t\in X:\\ \phi(x_t)=z_t}}} \mathsmaller{\frac{T(x_{t+1}|a_t,x_t) B_{\phi,t}^{\pi (k-1)}\left(x_t\middle|z_t,\{a_{t-i},z_{t-i}\}_{i=1}^{k-1}\right)}{T_{\phi,t}^{\pi(k-1)}\left(z_{t+1} \middle| \{a_{t-i}, z_{t-i}\}_{i=0}^{k-1}\right)}}\\
		&\Rightarrow & B_{\phi,t}^{\pi(k+1)}\left(x_{t+1}\middle|z_{t+1},\{a_{t-i},z_{t-i}\}_{i=0}^{k}\right) &= B_{\phi,t}^{\pi(k)}\left(x_{t+1}\middle|z_{t+1},\{a_{t-i},z_{t-i}\}_{i=0}^{k-1}\right).
	\end{align*}
\end{Proof}

We now summarize the proof of the main theorem. We begin by showing that the belief distributions $B_{\phi,t}^{\pi(k)}$ and $B_{\phi,t}^{\pi (k-1)}$ must be equal for $k=1$, and use Lemma \ref{lemma:B_step_to_TR_step} to prove the base case of the theorem. Then we use Lemma \ref{lemma:B_step_to_B_step} to prove that the theorem holds in general.

\begin{Proof}[of Theorem \ref{thm:markov-conditions}]\hspace{10px}

	\textbf{Base case.} For each $\pi\in \Pi_\phi$, let $B_{\phi,t}^\pi$ be defined via \eqref{eqn:B_phi}. Then, starting from the Density Ratio condition, for any $z_{t+1}, z_t \in Z$, and  $x_{t+1} \in X$ such that $\phi(x_{t+1})=z_{t+1}$, and any action $a_{t-1}\in A$:
	\begin{align*}
	    && \frac{P_{\phi,t}^\pi(z_{t}|z_{t-1})} {P_{\phi,t}^\pi(z_{t})} &=\ \frac{P_t^\pi(x_{t}|{z}_{t-1})}{P_t^\pi(x_{t})}\\
		&\Rightarrow & \frac{P_{\phi,t}^\pi(z_{t}|z_{t-1})} {P_{\phi,t}^\pi(z_{t})} &=\ \sum_{x_{t-1}\in X}{\frac{P_t^\pi(x_{t}|{x}_{t-1})}{P_t^\pi(x_{t})}}B_{\phi,t}^\pi({x}_{t-1}|z_{t-1})\\
		&\Rightarrow & \frac{P_t^\pi(x_{t})} {P_{\phi,t}^\pi(z_{t})} &=\ \sum_{x_{t-1}\in X}\mathsmaller{\frac{P_t^\pi(x_{t}|{x}_{t-1})B_{\phi,t}^\pi({x}_{t-1}|z_{t-1})}{P_{\phi,t}^\pi(z_{t}|z_{t-1})}\cdot \frac{I_t^\pi(a_{t-1}|x_{t},{x}_{t-1})\pi_{\phi,t}(a_{t-1}|z_{t-1})}{I_{\phi,t}^\pi(a_{t-1}|z_{t},z_{t-1})\pi_t(a_{t-1}|{x}_{t-1})}}\\
		&\Rightarrow & \frac{\ind[\phi(x_{t})=z_{t}]\ P_t^\pi(x_{t})} {\sum_{\tilde{x}_{t}\in z_{t}}P_t^\pi(\tilde{x}_{t})} &=\ \ind[\phi(x_{t})=z_{t}]\sum_{x_{t-1}\in X}\frac{T(x_{t}|a_{t-1},{x}_{t-1}) B_{\phi,t}^\pi({x}_{t-1}|z_{t-1})}{T_{\phi,t}^\pi(z_{t}|a_{t-1},z_{t-1})}\\
%
		%
		&\Rightarrow & B_{\phi,t}^\pi(x_{t}|z_{t}) &=\ \frac{\ind[\phi(x_{t})=z_{t}]\ \sum_{x_{t-1}\in X}T_{\phi}(x_{t}|a_{t-1},{x}_{t-1}) B_{\phi,t}^\pi({x}_{t-1}|z_{t-1})}{\sum_{\tilde{x}_{t}\in z_{t}}\sum_{\tilde{x}_{t-1}\in z_{t-1}} T(\tilde{x}_{t}|a_{t-1},\tilde{x}_{t-1}) B_{\phi,t}^\pi(\tilde{x}_{t-1}|z_{t-1})}\\
		&\Rightarrow & B_{\phi,t}^{\pi(0)}(x_{t}|z_{t}) &=\ B_{\phi,t}^{\pi(1)}(x_{t}|z_{t}, a_{t-1},z_{t-1}) \taghere \label{eqn:base_case_B}
	\end{align*}

	Here \eqref{eqn:base_case_B} satisfies the conditions of Lemma \ref{lemma:B_step_to_TR_step} (with $k=1$), therefore, for all $a_t \in A$:
	\begin{align*}
		T_{\phi,t}^{\pi(0)}(z_{t+1}|a_t,z_t) &= T_{\phi,t}^{\pi(1)}\left(z_{t+1}\middle|a_t,z_t,a_{t-1},z_{t-1}\right)\\
		\text{and} \quad R_{\phi,t}^{\pi(0)}(z_{t+1},a_t,z_t) &= R_{\phi,t}^{\pi(1)}(z_{t+1},a_t,z_t,a_{t-1},z_{t-1})
	\end{align*}
	This proves the theorem for $k=1$.

	\textbf{Induction on $k$.}
	Note that \eqref{eqn:base_case_B} also allows us to apply Lemma \ref{lemma:B_step_to_B_step}. Therefore, by induction on $k$:
	\[ B_{\phi,t}^{\pi(k+1)}\left(x_{t+1}\middle|z_{t+1},\{a_{t-i},z_{t-i}\}_{i=0}^{k}\right) = B_{\phi,t}^\pi(x_{t+1}|z_{t+1})\quad \forall\ k \ge 1\taghere \label{eqn:B_markov}\]
\end{Proof}

When \eqref{eqn:B_markov} holds, we informally say that the belief distribution is Markov.

\begin{Proof}[of Corollary \ref{corollary:markov-B-implies-MDP}] Follows directly from Definition \ref{def:abstract-markov} and Lemma \ref{lemma:B_step_to_TR_step} via induction on $k$.
\end{Proof}

This first corollary says that a Markov abstraction implies a Markov abstract state representation. The next one says that, if a belief distribution is non-Markov over some horizon $n$, it must also be non-Markov when conditioning on a single additional timestep.

\begin{corollary}
\label{corollary:B1_non_markov}
	If there exists some $n \ge 1$ such that $B_{\phi,t}^{\pi(n)} \ne B_{\phi,t}^\pi$, then $B_{\phi,t}^{\pi(1)} \ne B_{\phi,t}^\pi$.
\end{corollary}

\begin{Proof}
	Suppose such an $n$ exists, and assume for the sake of contradiction that $B_{\phi,t}^{\pi(1)} = B_{\phi,t}^\pi$. Then by Lemma \ref{lemma:B_step_to_B_step}, $B_{\phi,t}^{\pi(k)} = B_{\phi,t}^\pi$ for all $k \ge 1$. However this is impossible, since we know there exists some $n \ge 1$ such that $B_{\phi,t}^{\pi(n)} \ne B_{\phi,t}^\pi$. Therefore $B_{\phi,t}^{\pi(1)} \ne B_{\phi,t}^\pi$.
\end{Proof}


\subsection{Inverse Model Implies Density Ratio}
\label{appendix:inverse-implies-ratio}

As discussed in Section \ref{sec:counterexample}, the Inverse Model condition is not sufficient to ensure a Markov abstraction. In fact, what is missing is precisely the Density Ratio condition. Theorem \ref{thm:markov-conditions} already states that, given the Inverse Model condition, the Density Ratio condition is sufficient for an abstraction to be Markov over its policy class; the following theorem states that it is also necessary.

\begin{theorem}
\label{thm:inverse-implies-ratio}
If $\phi: X\rightarrow Z$ is a Markov abstraction of MDP $M = (X, A, R, T, \gamma)$ for any policy in the policy class $\Pi_\phi$, and the Inverse Model condition of Theorem \ref{thm:markov-conditions} holds for every timestep $t$, then the Density Ratio condition also holds for every timestep $t$.
\end{theorem}

\begin{Proof}\hspace{10px}

    Since $\phi$ is a Markov abstraction, equation \eqref{eqn:B_markov} holds for any $k \ge 1$. Fixing $k=1$, we obtain:
    \begin{align*}
        B_{\phi,t}^{\pi(1)}(x'|z', a, z) &= B_{\phi,t}^{\pi(0)}(x'|z') \\
        \frac{\ind[\phi(x')=z'] \sum_{\tilde x \in X} T(x'|a,\tilde x) B^\pi_{\phi,t}(\tilde x| z)}{\sum_{\tilde x' \in X: \phi(x') = z'} \sum_{\tilde x \in X: \phi(x)=z} T(\tilde x'| a, \tilde x) B^\pi_{\phi,t}(\tilde x| z)} &= \frac{\ind[\phi(x')=z'] P^\pi_{t}(x')}{\sum_{\tilde x'\in X: \phi(\tilde x') = z'} P^\pi_{t}(\tilde x')}\\
        \frac{\sum_{\tilde x \in X} T(x'|a,\tilde x) B^\pi_{\phi,t}( \tilde x| z)}{T^\pi_{\phi,t}(z'| a, z)} &= \frac{P^\pi_{t}(x')}{P^\pi_{\phi,t}(z')}\\
        \frac{\sum_{\tilde x \in X} T(x'|a,\tilde x) B^\pi_{\phi,t}( \tilde x| z)}{P^\pi_{t}(x')} &= \frac{T^\pi_{\phi,t}(z'| a, z)}{P^\pi_{\phi,t}(z')}\\
        \mathlarger{\sum}_{\tilde x \in X}\ \frac{I^\pi_t(a|x',\tilde x) P^\pi_{t}(x'|\tilde x) B^\pi_{\phi,t}(\tilde x| z)}{P^\pi_{t}(x') \pi(a|\tilde x)} &= \frac{I^\pi_{\phi,t}(a|z', z) P^\pi_{\phi,t}(z'|z)}{P^\pi_{\phi,t}(z')\pi(a|z)}\\
        \frac{\sum_{\tilde x \in X}I^\pi_t(a|x',\tilde x) P^\pi_{t}(x'|\tilde x)B^\pi_{\phi,t}(\tilde x| z)}{P^\pi_{t}(x')} &= \frac{I^\pi_{\phi,t}(a|z', z) P^\pi_{\phi,t}(z'|z)}{P^\pi_{\phi,t}(z')} \taghere \label{eqn:inverse-ratio-product}\\
        \intertext{Here we apply the Inverse Model condition, namely that $I^\pi_t(a|x',x) = I^\pi_{\phi,t}(a|z',z)$ for all $z,z'\in Z$; $x,x'\in X$, such that $\phi(x')=z'$ and $\phi(x)=z$.}
        \frac{\sum_{\tilde x \in X} P^\pi_{t}(x'|\tilde x) B^\pi_{\phi,t}(\tilde x| z)}{P^\pi_{t}(x') }  &= \frac{ P^\pi_{\phi,t}(z'|z)}{P^\pi_{\phi,t}(z')}\\
        \frac{P^\pi_{t}(x'|z)}{P^\pi_{t}(x') } &= \frac{ P^\pi_{\phi,t}(z'|z)}{P^\pi_{\phi,t}(z')} \tag{Density Ratio}
    \end{align*}
\end{Proof}
It may appear that Theorems \ref{thm:markov-conditions} and \ref{thm:inverse-implies-ratio} together imply that the Inverse Model and Density Ratio conditions are necessary and sufficient for an abstraction to be Markov over its policy class; however, this is not quite true. Both conditions, taken together, are sufficient for an abstraction to be Markov, and, \emph{given the Inverse Model condition}, the Density Ratio condition is necessary. Examining equation \eqref{eqn:inverse-ratio-product}, we see that, had we instead assumed the Density Ratio condition for Theorem \ref{thm:inverse-implies-ratio} (rather than the Inverse Model condition), we would not recover $I^\pi_t(a|x',x) = I^\pi_{\phi,t}(a|z',z)$, but rather $\sum_{\tilde x \in X} I^\pi_t(a|x',\tilde x) B^\pi_{\phi,t}(\tilde x| z) = I^\pi_{\phi,t}(a|z',z)$. That is, the Inverse Model condition would only be guaranteed to hold in expectation, but not for arbitrary $x \in X$.

\newpage
\section{Derivation of Density Ratio Objective}
\label{appendix:ratio-derivation}

Our Density Ratio objective in Section \ref{sec:training} is based on the following derivation, adapted from \citet{tiao2017simple}.

Suppose we have a dataset consisting of samples $X_c = \{{x'_c}^{(i)}\}_{i=1}^{n_c}$ drawn from conditional distribution $\Pr(x'|x)$, and samples $X_m = \{{x'_m}^{(j)}\}_{j=1}^{n_m}$ drawn from marginal distribution $\Pr(x')$.
We assign label $y = 1$ to samples from $X_c$ and $y=0$ to samples from $X_m$, and our goal is to predict the label associated with each sample.
To construct an estimator, we rename the two distributions $p(x'|y=1) := \Pr(x'|z)$ and $p(x'|y=0) := \Pr(x')$ and rewrite the density ratio $\delta(x') := \frac{\Pr(x'|x)}{\Pr(x')}$ as follows:
\[
\begin{aligned}
    \delta(x') =
    \mathsmaller{\frac{p(x'|y=1)}{p(x'|y=0)}} =
    \mathsmaller{\frac{p(y=1|x')p(x')}{p(y=1)}\frac{p(y=0)}{p(y=0|x')p(x')}} = \mathsmaller{\frac{n_m}{(n_m+n_c)}\frac{(n_m+n_c)}{n_c}\frac{p(y=1|x')}{p(y=0|x')}} =
    \mathsmaller{\frac{n_m}{n_c}\frac{p(y=1|x')}{1-p(y=1|x')}}.
\end{aligned}\taghere \label{eqn:delta} \]

When $n_c = n_m = N$, which is the case for our implementation, the leading fraction can be ignored. To estimate $\delta(x')$, we can simply train a classifier $g(x',x;\theta_g)$ to approximate $p(y=1|x')$ and then substitute $g$ for $p(y=1|x')$ in \eqref{eqn:delta}.

However, we need not estimate $\delta(x')$ to satisfy the Density Ratio condition; we need only ensure $\delta_\phi(z') = \E_{B_\phi}[\delta(x')]$. We therefore repeat the derivation for abstract states and obtain $\delta_\phi(z') := \frac{\Pr(z'|z)}{\Pr(z')} = \frac{n_m}{n_c}\frac{q(y=1|z')}{1-q(y=1|z')}$, where $q$ is our renamed distribution, and modify our classifier $g$ to accept abstract states instead of ground states. The labels are the same regardless of whether we use ground or abstract state pairs, so training will cause $g$ to approach $p$ and $q$ simultaneously, thus driving $\delta_\phi(z') \rightarrow \delta(x')$ and satisfying the Density Ratio condition.

\newpage

\section{Markov State Abstractions and Kinematic Inseparability}
\label{appendix:KI}
As discussed in Section \ref{sec:related-work}, the notion of kinematic inseparability \citep{misra2020kinematic} is closely related to Markov abstraction. Recall that two states $x_1'$ and $x_2'$ are defined to be kinematically inseparable if $\Pr(x,a|x'_1) = \Pr(x,a|x'_2)$ and $T(x''|a,x_1')=T(x''|a,x_2')$ (which the authors call ``backwards'' and ``forwards'' KI, respectively). \citet{misra2020kinematic} define kinematic inseparability abstractions over the set of all possible ``roll-in'' distributions $u(x,a)$ supported on $X\times A$, and technically, the backwards KI probabilities $\Pr(x,a|x')$ depend on $u$. However, to support choosing a policy class, we can just as easily define $u$ in terms of a policy: $u(x,a) := \pi(a|x)P^\pi_t(x)$. This formulation leads to:
\[ P^\pi_t(x,a|x') := \frac{T(x'|a,x)\pi(a|x)P^\pi_t(x)} {\sum_{\tilde x \in X, \tilde a \in A} T(x'|\tilde a,\tilde x)\pi(\tilde a|\tilde x)P^\pi_t(\tilde x)}.\]

\begin{definition}
An abstraction $\phi: X \rightarrow Z$ is a \emph{kinematic inseparability abstraction} of MDP $M=(X,A,R,T,\gamma)$ over policy class $\Pi_C$, if for all policies $\pi \in \Pi_C$, and all $a\in A; x, x'_1, x'_2, x'' \in X$ such that $\phi(x'_1) = \phi(x'_2)$; $P^\pi_t(x,a|x'_1) = P^\pi_t(x,a|x'_2)$ and $T(x''|a,x_1')=T(x''|a,x_2')$.
\end{definition}

Similarly, we can define forward---or backward---KI abstractions where only $T(x''|a,x'_1) = T(x''|a,x'_2)$---or respectively, $P^\pi_t(x,a|x'_1)=P^\pi_t(x,a|x'_2)$---is guaranteed to hold. A KI abstraction is one that is both forward KI and backward KI.

The KI conditions are slightly stronger conditions than those of Theorem \ref{thm:markov-conditions}, as the following example demonstrates.

\subsection{Example MDP}
The figure below modifies the transition dynamics of the MDP in Section \ref{sec:background}, such that the action $a_1$ has the same effect everywhere: to transition to either central state, $x_1$ or $x_2$, with equal probability.

\begin{figure}[!h]
  \centering
  \includegraphics[width=.6\textwidth]{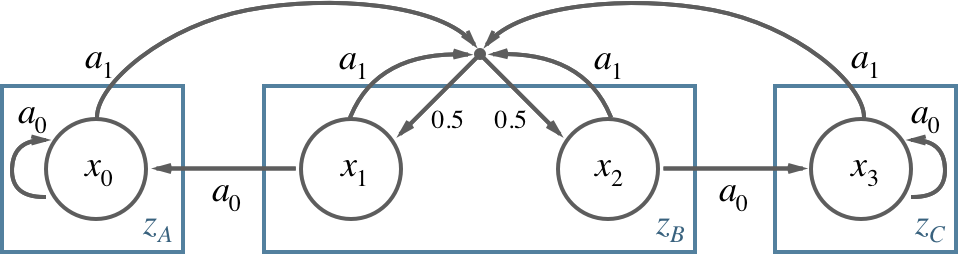}
  \captionof{figure}{An MDP and a Markov abstraction that is not a KI abstraction.}
  \label{fig:ki-counterexample}
\end{figure}

By the same reasoning as in Section \ref{sec:counterexample}, the Inverse Model condition holds here, but now, due to the shared transition dynamics of action $a_1$, the Density Ratio condition holds as well, for any policy in $\Pi_\phi$. We can apply Theorem \ref{thm:markov-conditions} to see that the abstraction is Markov, or we can simply observe that conditioning the belief distribution $B^\pi_{\phi,t}(x|z_B)$ on additional history has no effect, since any possible trajectory ending in $z_B$ leads to the same 50--50 distribution over ground states $x_1$ and $x_2$. Either way, $\phi$ is Markov by Definition \ref{def:abstract-markov}.

This abstraction also happens to satisfy the backwards KI condition, since $P^\pi_t(x,a|x'_1) = P^\pi_t(x,a|x'_2)$ for any $(x,a)$ pair and any policy. However, clearly $T(x'|a_0,x_1)\ne T(x'|a_0,x_2)$, and therefore the forwards KI condition does not hold and this is not a KI abstraction.

This example shows that the Markov conditions essentially take the stance that because $\pi$ is restricted to the policy class $\Pi_\phi$, knowing the difference between $x_1$ and $x_2$ doesn't help, because $\pi$ must have the same behavior for both ground states. By contrast, the KI conditions take the stance that because $x_1$ and $x_2$ have different dynamics, the agent may wish to change its behavior based on which state it sees, so it ought to choose an abstraction that does not limit decision making in that respect.


\subsection{``Strongly Markov'' Implies KI}
In Section \ref{sec:training}, we mentioned that the Density Ratio training objective was stronger than necessary to ensure the corresponding condition of Theorem \ref{thm:markov-conditions}. Instead of encoding the condition $\frac{\Pr(x'|z)}{\Pr(x')} = \frac{\Pr(z'|z)}{\Pr(z')}$, we discussed how the contrastive training procedure actually encodes the stronger condition $\frac{\Pr(x'|x)}{\Pr(x')} = \frac{\Pr(z'|z)}{\Pr(z')}$ that holds for each ground state $x$ individually, rather than just in expectation. Let us call the latter condition the Strong Density Ratio condition, and call its combination with the Inverse Model condition the Strong Markov conditions.

Clearly the Strong Markov conditions imply the original Markov conditions, and, as the following theorem shows, they also imply the KI conditions.

\begin{theorem}
If $\phi: X \rightarrow Z$ is an abstraction of MDP $M=(X,A,R,T,\gamma)$ such that for any policy $\pi$ in the policy class $\Pi_\phi$, both the Inverse Model condition of Theorem 1 and the Strong Density Ratio condition---i.e. $\frac{P^\pi_t(x'|x)}{P^\pi_t(x')} = \frac{P^\pi_{\phi,t}(z'|z)}{P^\pi_{\phi,t}(z')}$, for all $z,z'\in Z$; $x,x'\in X$ such that $\phi(x)=z$ and $\phi(x')=z'$---hold for every timestep $t$, then $\phi$ is a kinematic inseparability abstraction .
\end{theorem}

\begin{Proof}\hspace{10px}

    Starting from the Inverse Model condition, we have $I^\pi_t(a|x',x) = I^\pi_{\phi,t}(a|z',z)$ for all $z,z' \in Z$; $x,x'\in X$ such that $\phi(x)=z$ and $\phi(x')=z'$. Independently varying either $x' \in \phi^{-1}(z')$ or $x \in \phi^{-1}(z)$, we obtain the following:
    \begin{align}
        &\text{[Vary $x'$]} \rightarrow \quad I^\pi_t(a|x_1',x) = I^\pi_t(a|x_2',x) \label{eqn:left-inv}\\
        &\text{[Vary $x$]} \rightarrow \quad I^\pi_t(a|x',x_1) = I^\pi_t(a|x',x_2) \label{eqn:right-inv}
    \end{align}
    Similarly, if we start from the Strong Density Ratio condition, we obtain:
    \begin{align}
        &\text{[Vary $x'$]} \rightarrow \quad \frac{P^\pi_t(x_1'|x)}{P^\pi_t(x_1')} = \frac{P^\pi_t(x_2'|x)}{P^\pi_t(x_2')} \label{eqn:left-strong-ratio}\\
        &\text{[Vary $x$]} \rightarrow \quad \frac{P^\pi_t(x'|x_1)}{P^\pi_t(x')} = \frac{P^\pi_t(x'|x_2)}{P^\pi_t(x')} \label{eqn:right-strong-ratio}
    \end{align}
    If we apply Bayes' theorem to \eqref{eqn:left-strong-ratio}, we can cancel terms in the result, and also in \eqref{eqn:right-strong-ratio}, to obtain:
    \begin{align}
        \quad P^\pi_t(x|x_1') &= P^\pi_t(x|x_2') \label{eqn:left-strong-next}\\
        \text{and}\quad P^\pi_t(x'|x_1) &= P^\pi_t(x'|x_2). \label{eqn:right-strong-next}
    \end{align}
    Combining \eqref{eqn:left-inv} with \eqref{eqn:left-strong-next}, we obtain the backwards KI condition:
    \begin{align*}
        I^\pi_t(a|x_1',x)P^\pi_t(x|x_1') &= I^\pi_t(a|x_2',x)P^\pi_t(x|x_2') \\
        P^\pi_t(x,a|x_1') &= P^\pi_t(x,a|x_2') \tag{Backwards KI}
    \end{align*}
    Similarly, we can combine \eqref{eqn:right-inv} with \eqref{eqn:right-strong-next} to obtain the forwards KI condition:
    \begin{align*}
        I^\pi_t(a|x',x_1)P^\pi_t(x'|x_1) &= I^\pi_t(a|x',x_2)P^\pi_t(x'|x_2) \\
        T(x'|a,x_1)\pi(a|x_1) &= T(x'|a,x_2)\pi(a|x_2)\\
        T(x'|a,x_1) &= T(x'|a,x_2) \tag{Forwards KI}
    \end{align*}
\end{Proof}

Thus, we see that the training objectives in Section \ref{sec:training} encourage learning a kinematic inseparability abstraction in addition to a Markov abstraction. This helps avoid representation collapse by ensuring that we do not group together any states for which a meaningful kinematic distinction can be made.

\newpage

\section{Implementation Details for Visual Gridworld}
\label{appendix:gridworld}
The visual gridworld is a $6 \times 6$ grid with four discrete actions: up, down, left, and right. Observed states are generated by converting the agent's $(x,y)$ position to a one-hot image representation (see Figure \ref{fig:main-objective}, left). The image displays each position in the $6 \times 6$ grid as a 3px-by-3px patch, inside of which we light up one pixel (in the center) and then smooth it using a truncated Gaussian kernel. This results in an $18 \times 18$ image (where 3px-by-3px grid cells are equidistant), to which we then add per-pixel noise from another truncated Gaussian. During pretraining, there are no rewards or terminal states. During training, for each random seed, a single state is designated to be the goal state, and the agent receives $-1$ reward per timestep until it reaches the goal state, at which point a new episode begins, with the agent in a random non-goal location.

\subsection{Computing Resources}
To build the figures in the paper, we pretrained 5 different abstractions, and trained 7 different agents, each with 300 seeds. Each 3000-step pretraining run takes about 1 minute, and each training run takes about 30 seconds, on a 2016 MacBook Pro 2GHz i5 with no GPU, for a total of about 42 compute hours. We ran these jobs on a computing cluster with comparable processors or better.

\subsection{Network Architectures}

\begin{lstlisting}[language={},basicstyle=\footnotesize\ttfamily,breaklines=true]
FeatureNet(
  (phi): Encoder(
    (0): Reshape(-1, 252)
    (1): Linear(in_features=252, out_features=32, bias=True)
    (2): Tanh()
    (3): Linear(in_features=32, out_features=2, bias=True)
    (4): Tanh()
  )
  (inv_model): InverseNet(
    (0): Linear(in_features=4, out_features=32, bias=True)
    (1): Tanh()
    (2): Linear(in_features=32, out_features=4, bias=True)
  )
  (contr_model): ContrastiveNet(
    (0): Linear(in_features=4, out_features=32, bias=True)
    (1): Tanh()
    (2): Linear(in_features=32, out_features=1, bias=True)
    (3): Sigmoid()
  )
)
QNet(
  (0): Linear(in_features=2, out_features=32, bias=True)
  (1): ReLU()
  (2): Linear(in_features=32, out_features=4, bias=True)
)
AutoEncoder / PixelPredictor(
  (phi): Encoder(
    (0): Reshape(-1, 252)
    (1): Linear(in_features=252, out_features=32, bias=True)
    (2): Tanh()
    (3): Linear(in_features=32, out_features=2, bias=True)
    (4): Tanh()
  )
  (phi_inverse): Decoder(
    (0): Linear(in_features=2, out_features=32, bias=True)
    (1): Tanh()
    (2): Linear(in_features=32, out_features=252, bias=True)
    (3): Tanh()
    (4): Reshape(-1, 21, 12)
  )
  MSELoss()
)
\end{lstlisting}

\subsection{Hyperparameters}
We tuned the DQN hyperparameters until it learned effectively with expert features (i.e. ground-truth $(x,y)$ position), then we left the DQN hyperparameters fixed while tuning pretraining hyperparameters. For pretraining, we considered 3000 and 30,000 gradient updates (see Appendix \ref{appendix:increased-pretraining}), and batch sizes within \{512, 1024, 2048\}. We found that the higher batch size was helpful for stabilizing the offline representations. We also did some informal experiments with latent dimensionality above 2, such as 3 or 10, which produced similar results: representations were still Markov, but harder to interpret. We use 2 dimensions in the paper for ease of visualization. We did not tune the loss coefficients, but we include ablations where either $\alpha$ or $\beta$ is set to zero.

\begin{table}[htbp]
\parbox{.45\linewidth}{
    \centering
    \small
    \begin{tabular}{lc}
    \toprule
    \textbf{Hyperparameter} & \textbf{Value} \\
    \midrule
    Number of seeds & 300 \\
    Optimizer & Adam \\
    Learning rate & 0.003 \\
    Batch size & 2048 \\
    Gradient updates & 3000 \\
    Latent dimensions & 2 \\
    Number of conditional samples, $n_c$ & 1 \\
    Number of marginal samples, $n_m$ & 1 \\
    Loss coefficients\\
    \hspace{10px}$\mathcal{L}_{\text{Inverse}}$ $(\alpha)$ & 1.0 \\
    \hspace{10px}$\mathcal{L}_{\text{Contrastive}}$ $(\beta)$ & 1.0 \\
    \hspace{10px}$\mathcal{L}_{\text{Smoothness}}$ $(\eta)$ & 0.0 \\
    \bottomrule
    \end{tabular}
    \caption{Pretraining hyperparameters}
    \label{tab:hyperparams-pretraining}
}
\hfill
\parbox{.45\linewidth}{
    \centering
    \small
    \begin{tabular}{lc}
    \toprule
    \textbf{Hyperparameter} & \textbf{Value} \\
    \midrule
    Number of seeds & 300 \\
    Number of episodes & 100 \\
    Maximum steps per episode & 1000 \\
    Optimizer & Adam \\
    Learning rate & 0.003 \\
    Batch size & 16 \\
    Discount factor, $\gamma$ & 0.9 \\
    Starting exploration probability, $\epsilon_0$ & 1.0 \\
    Final exploration probability, $\epsilon$ & 0.05 \\
    Epsilon decay period & 2500 \\
    Replay buffer size & 10000\\
    Initialization steps & 500\\
    Target network copy period & 50\\
    \bottomrule
    \end{tabular}
    \caption{DQN hyperparameters}
    \label{tab:hyperparams-rl}
}
\end{table}


\newpage

\section{Implementation Details for DeepMind Control}
\label{appendix:deepmind-control}
We use the same RAD network architecture and code implementation as \citet{laskin2020reinforcement}, which we customized to add our Markov objective. We note that there was a discrepancy between the batch size in their code implementation (128) and what was reported in the original paper (512); we chose the former for our experiments.

The SAC (expert) results used the code implementation from \citet{pytorch_sac}.

The DBC, DeepMDP, CPC, and SAC-AE results are from \citet{zhang2021learning}, except for Ball\_in\_Cup, which they did not include in their experimental evaluation. We ran their DBC code independently (with the same settings they used) to produce our Ball\_in\_Cup results.

\subsection{Computing Resources}
To build the graph in Section \ref{sec:online}, we trained 4 agents on 6 domains, with 10 seeds each. Each training run (to 500,000 steps) takes between 24 and 36 hours (depending on the action repeat for that domain), on a machine with two Intel Xeon Gold 5122 vCPUs and shared access to one Nvidia 1080Ti GPU, for a total of approximately 7200 compute hours. We ran these jobs on a computing cluster with comparable hardware or better.

\subsection{Markov Network Architecture}

\begin{lstlisting}[language={},basicstyle=\footnotesize\ttfamily,breaklines=true]
MarkovHead(
  InverseModel(
    (body): Sequential(
      (0): Linear(in_features=100, out_features=1024, bias=True)
      (1): ReLU()
      (2): Linear(in_features=1024, out_features=1024, bias=True)
      (3): ReLU()
    )
    (mean_linear): Linear(in_features=1024, out_features=action_dim, bias=True)
    (log_std_linear): Linear(in_features=1024, out_features=action_dim, bias=True)
  )
  ContrastiveModel(
    (model): Sequential(
      (0): Linear(in_features=100, out_features=1024, bias=True)
      (1): ReLU()
      (2): Linear(in_features=1024, out_features=1, bias=True)
    )
  )
  BCEWithLogitsLoss()
)
\end{lstlisting}


\subsection{Hyperparameters}

When tuning our algorithm, we left all RAD hyperparameters fixed except init\_steps, which we increased to equal 10 episodes across all domains to provide adequate coverage for pretraining. We compensated for this change by adding (init\_steps - 1K) catchup learning steps to ensure both methods had the same number of RL updates. This means our method is at a slight disadvantage, since RAD can begin learning from reward information after just 1K steps, but our method must wait until after the first 10 episodes of uniform random exploration. Otherwise, we only considered changes to the Markov hyperparameters (see Table \ref{tab:hyperparams-Markov}). We set the Markov learning rate equal to the RAD learning rate for each domain (and additionally considered 5e-5 for cheetah only). We tuned the $\mathcal{L}_{Inv}$ loss coefficient within \{0.1, 1.0, 10.0, 30.0\}, and the $\mathcal{L}_{Smooth}$ loss coefficient within \{0, 10.0, 30.0\}. The other hyperparameters, including the network architecture, we did not change from their initial values.

{\centering
\begin{table}[htbp]
\parbox{.45\linewidth}{
    \centering
    \small
    \begin{tabular}{ll}
    \toprule
    \textbf{Hyperparameter} & \textbf{Value} \\
    \midrule
    Augmentation\\
    \hspace{10px} Walker & Crop\\
    \hspace{10px} Others & Translate\\
    Observation rendering & (100, 100)\\
    Crop size & (84, 84)\\
    Translate size & (108, 108)\\
    Replay buffer size & 100000\\
    Initial steps & 1000\\
    Stacked frames & 3\\
    Action repeat & 2; finger,\\
    & \hphantom{2;} walker\\
    & 8; cartpole \\
    & 4; others\\
    Hidden units (MLP) & 1024 \\
    Evaluation episodes & 10 \\
    Optimizer & Adam \\
    \hspace{10px}$(\beta_1, \beta_2) \rightarrow (\phi, \pi, Q)$ & (.9, .999) \\
    \hspace{10px}$(\beta_1, \beta_2) \rightarrow (\alpha)$ & (.5, .999) \\
    \hspace{10px}Learning rate $(\phi, \pi, Q)$ & 2e-4, cheetah\\
    & 1e-3, others\\
    \hspace{10px}Learning rate $(\alpha)$ & 1e-4 \\
    Batch size & 128 \\
    Q function EMA $\tau$ & 0.01 \\
    Critic target update freq & 2 \\
    Convolutional layers & 4 \\
    Number of filters & 32 \\
    Non-linearity & ReLU \\
    Encoder EMA $\tau$ & 0.05 \\
    Latent dimension & 50 \\
    Discount $\gamma$ & .99\\
    Initial temperature & 0.1 \\
    \bottomrule
    \end{tabular}
    \caption{RAD hyperparameters}
    \label{tab:hyperparams-RAD}
}
\hspace{1px}
\parbox{.5\linewidth}{
    \centering
    \small
    \begin{tabular}{lll}
    \toprule
    \textbf{Hyperparameter} & \textbf{Value} \\
    \midrule
    Pretraining steps & 100K\\
    Pretraining batch size & 512\\
    RAD init steps & \multicolumn{2}{l}{(20K / action\_repeat)}\\
    RAD catchup steps & \multicolumn{2}{l}{(init\_steps - 1K)}\\
    Other RAD parameters & unchanged \\

    Loss coefficients\\
    \hspace{10px}$\mathcal{L}_{Inv}$ & 30.0 & ball,\\
    && reacher\\
    & 1.0 & others\\
    \hspace{10px}$\mathcal{L}_{Ratio}$ & 1.0 \\
    \hspace{10px}$\mathcal{L}_{Smooth}$ & 30.0 & ball,\\
    && reacher,\\
    && cheetah\\
    & 10.0 & others\\
    Smoothness $d_0$ & 0.01 \\
    Conditional samples, $n_c$ & 128 \\
    Marginal samples, $n_m$ & 128 \\
    Optimizer & Adam\\
    \hspace{10px}$(\beta_1, \beta_2) \rightarrow$ (Markov) & (.9, .999)\\
    \hspace{10px}Learning rate & 2e-4 & cheetah\\
    & 1e-3 & others\\

    \bottomrule
    \end{tabular}
    \caption{Markov hyperparameters}
    \label{tab:hyperparams-Markov}
}
\end{table}
}

\clearpage

\newpage

\section{Additional Representation Visualizations}
\label{appendix:more-visualizations}

Here we visualize abstraction learning progress for the $6 \times 6$ visual gridworld domain for six random seeds. Each figure below displays selected frames (progressing from left to right) of a different abstraction learning method (top to bottom): $\mathcal{L}_{Markov}$; $\mathcal{L}_{Inv}$ only; $\mathcal{L}_{Ratio}$ only; autoencoder; pixel prediction. The networks are initialized identically for each random seed. Color denotes ground-truth $(x,y)$ position, which is not shown to the agent. These visualizations span 30,000 training steps (columns, left to right: after 1, 100, 200, 700, 3K, 10K, and 30K steps, respectively). In particular, the third column from the right shows the representations after 3000 steps, which we use for the results in the main text. We show additional learning curves for the final representations in Appendix \ref{appendix:increased-pretraining}.
\vspace{10px}

\begin{figure}[bh]
    \centering
    \begin{subfigure}{.48\textwidth}
        \includegraphics[width=.99\linewidth]{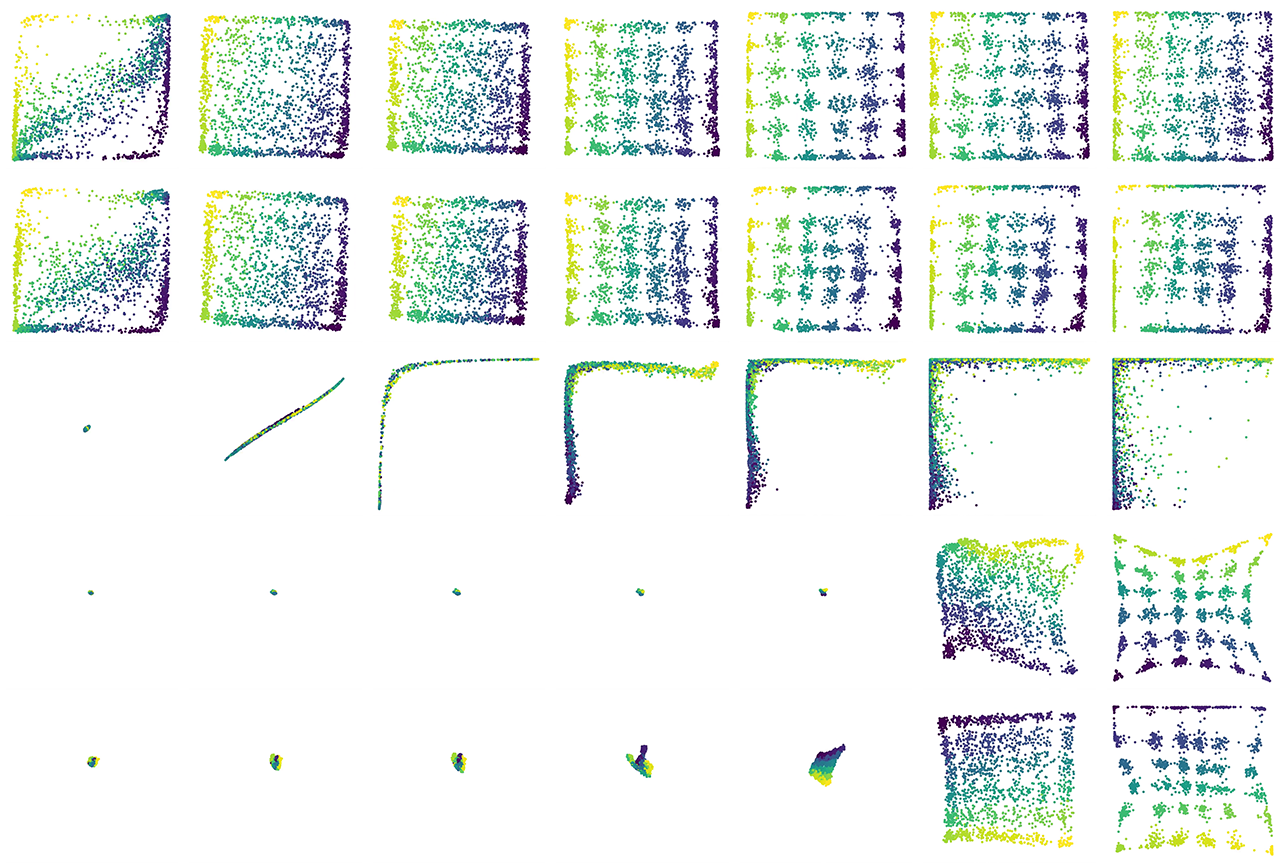}
        \caption*{Seed 1}
    \end{subfigure}
    \quad
    \begin{subfigure}{.48\textwidth}
        \includegraphics[width=.99\linewidth]{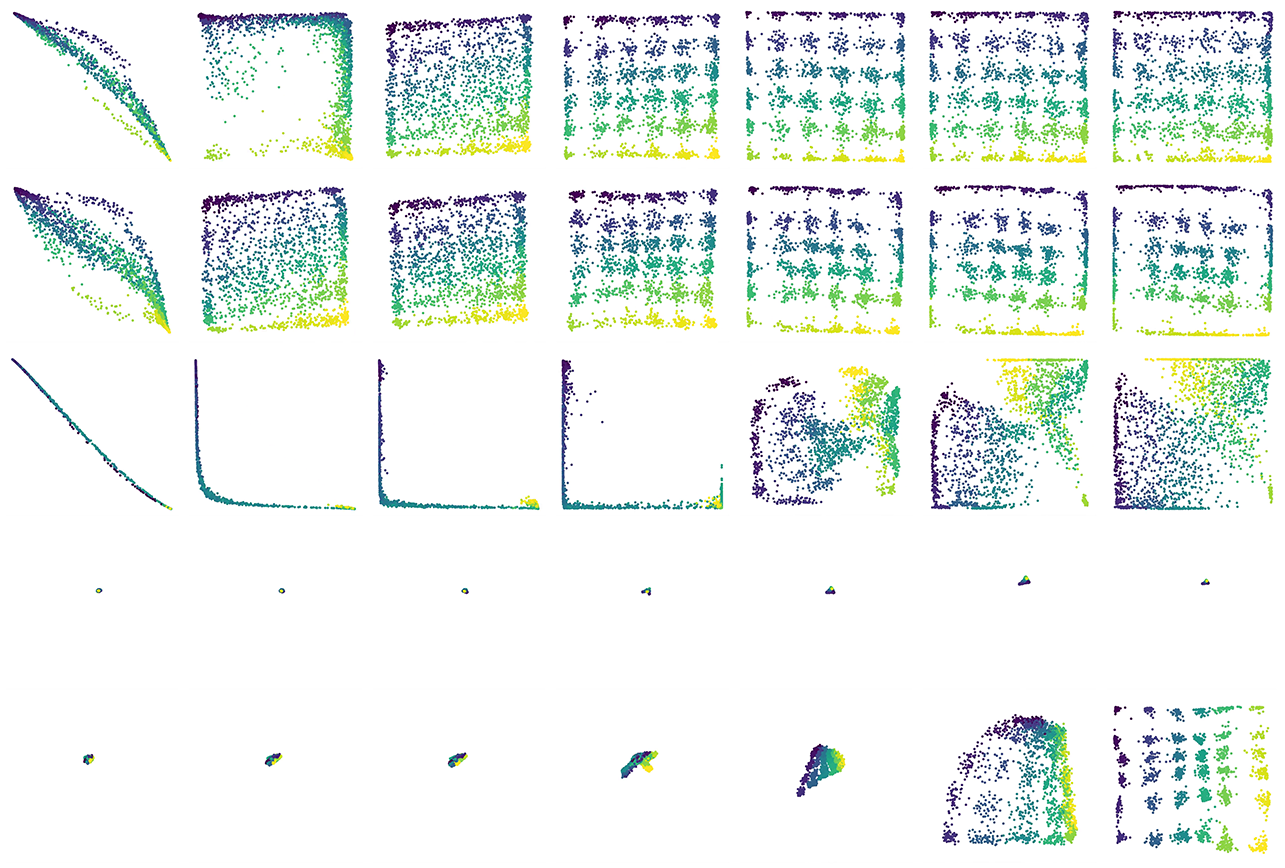}
        \caption*{Seed 2}
    \end{subfigure}\\
    \hspace{1px}\vspace{20px}
    \\
    \begin{subfigure}{.48\textwidth}
        \includegraphics[width=.99\linewidth]{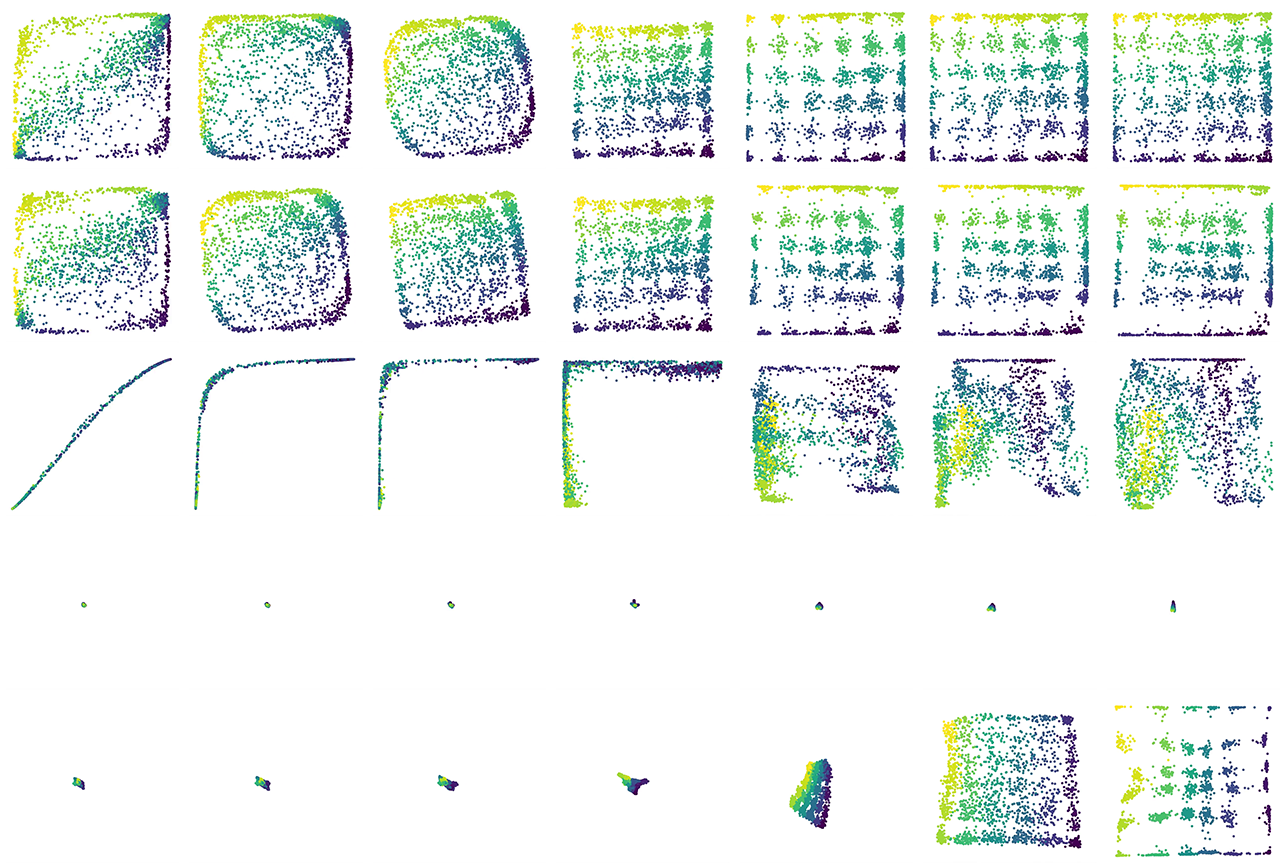}
        \caption*{Seed 3}
    \end{subfigure}
    \quad
    \begin{subfigure}{.48\textwidth}
        \includegraphics[width=.99\linewidth]{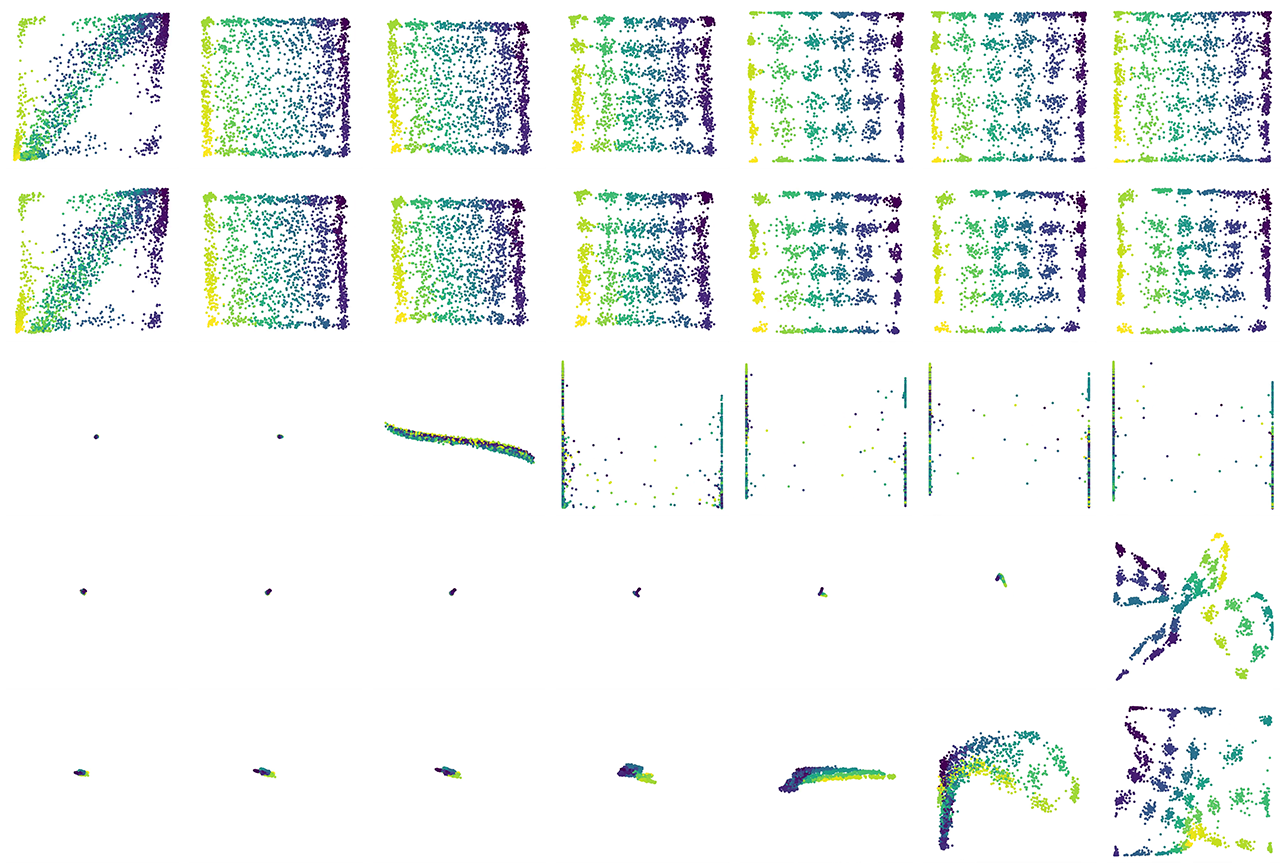}
        \caption*{Seed 4}
    \end{subfigure}\\
    \hspace{1px}\vspace{20px}
    \\
    \begin{subfigure}{.48\textwidth}
        \includegraphics[width=.99\linewidth]{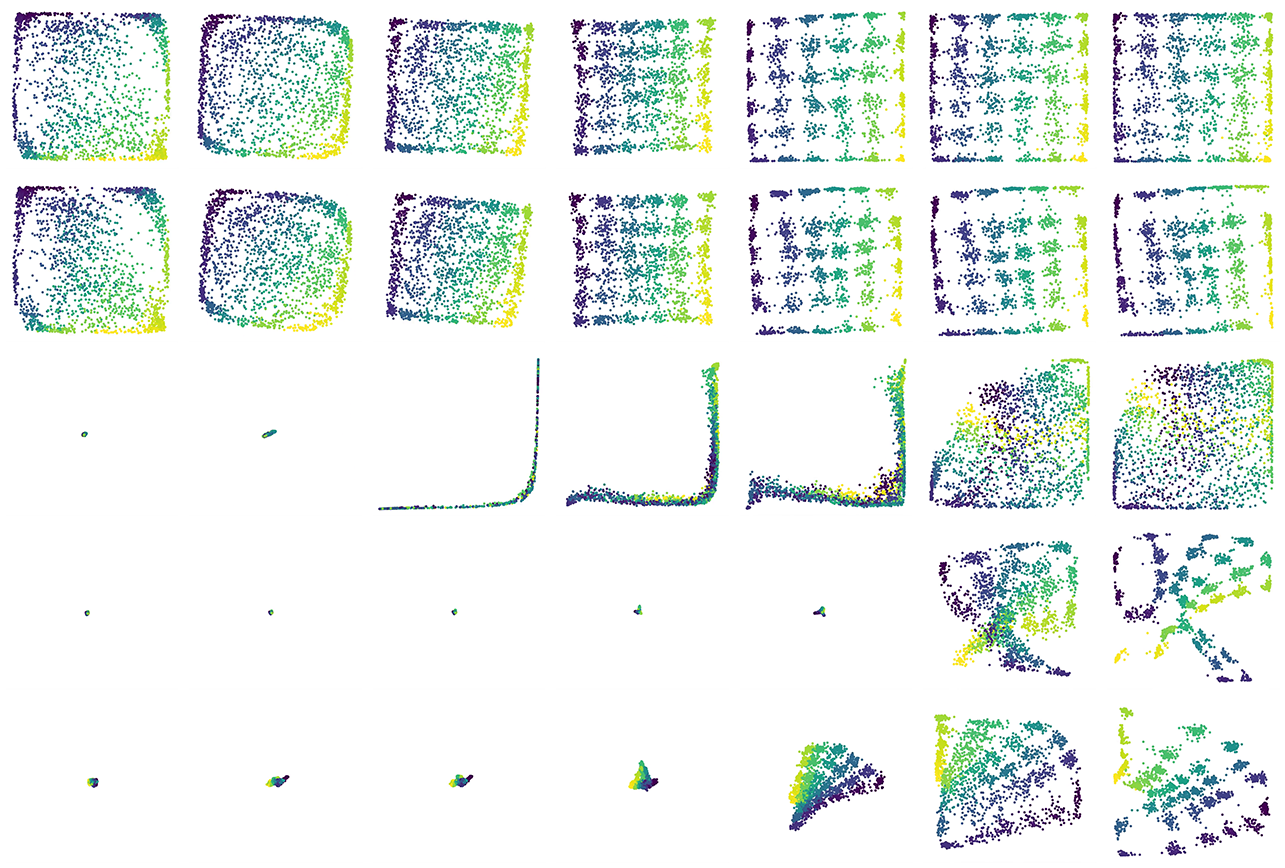}
        \caption*{Seed 5}
    \end{subfigure}\quad
    \begin{subfigure}{.48\textwidth}
        \includegraphics[width=.99\linewidth]{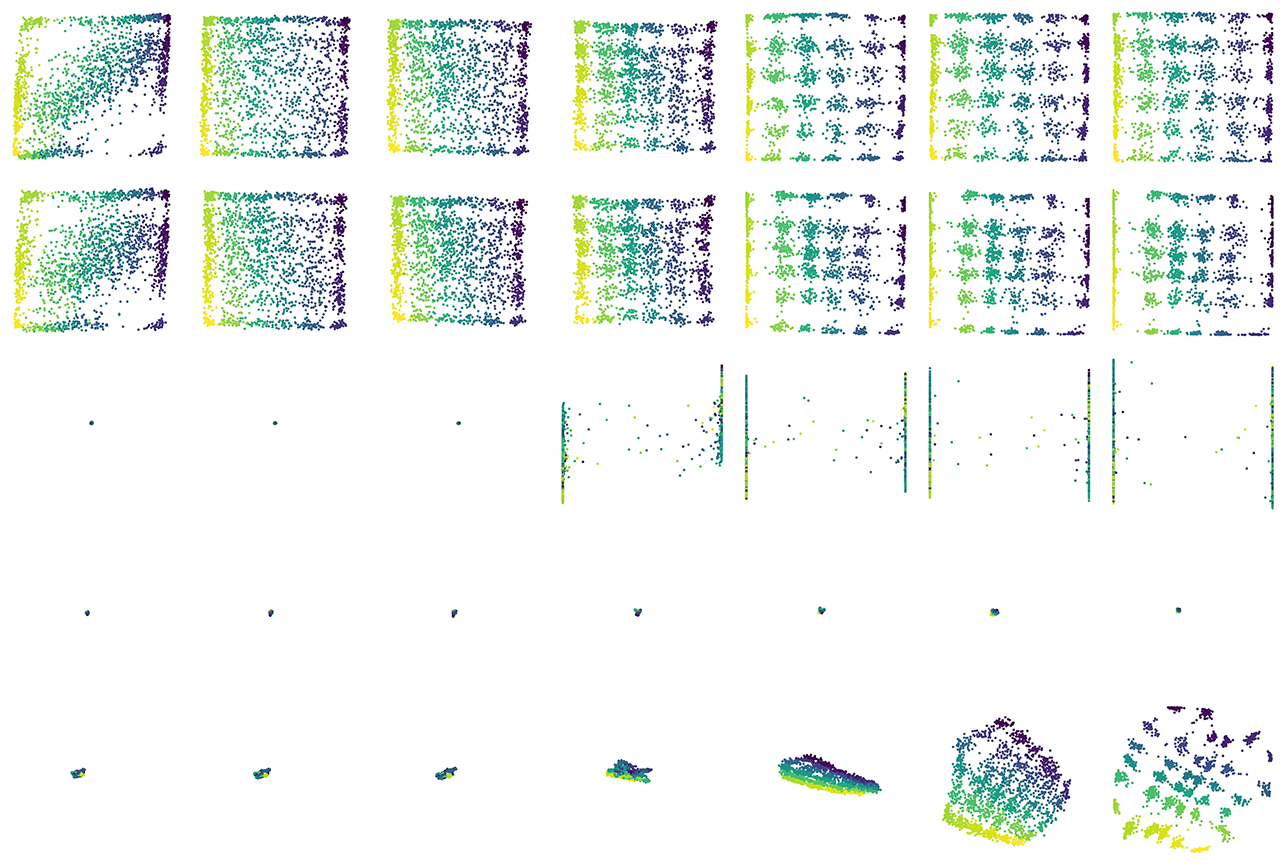}
        \caption*{Seed 6}
    \end{subfigure}
    \captionof{figure}{}
    \label{fig:rep-vis-tall}
\end{figure}

\newpage
\section{Gridworld Results for Increased Pretraining Time}
\label{appendix:increased-pretraining}
\begin{figure}[!htbp]
  \centering
  \includegraphics[width=.98\linewidth]{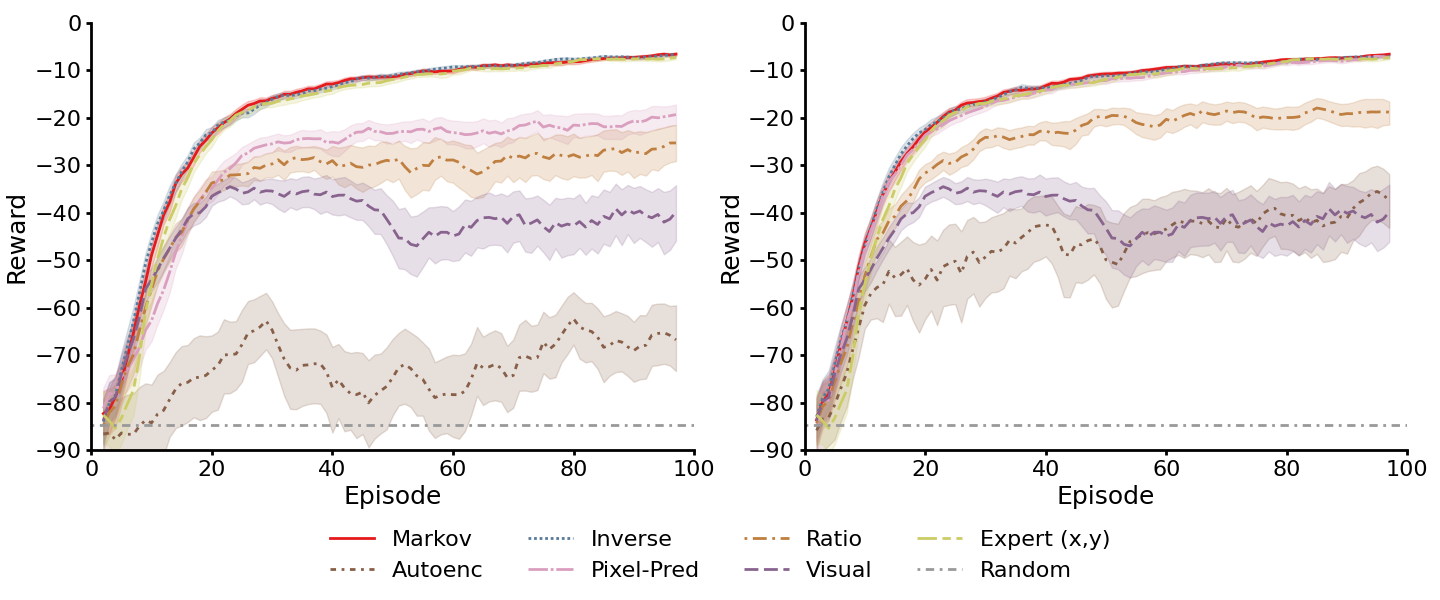}
  \captionof{figure}{Mean episode reward for the visual gridworld navigation task, using representations that were pretrained for 3,000 steps (left) versus 30,000 steps (right). Increased pretraining time improves the performance of $\phi_{Ratio}$, $\phi_{Autoenc}$ and $\phi_{PixelPred}$. (300 seeds; 5-point moving average; shaded regions denote 95\% confidence intervals.)}
  \label{fig:rep-eval-30k}
\end{figure}

Since some of the representations in Appendix \ref{appendix:more-visualizations} appeared not to have converged after just 3000 training steps, we investigated whether the subsequent learning performance would improve with more pretraining. We found that increasing the number of pretraining steps from 3000 to 30,000 improves the learning performance of $\phi_{Ratio}$ and $\phi_{Autoenc}$ and $\phi_{PixelPred}$ (see Figure \ref{fig:rep-eval-30k}), with the latter representation now matching the performance of $\phi_{Markov}$.

It is perhaps unsurprising that the pixel prediction model eventually recovers the performance of the Markov abstraction, because the pixel prediction task is a valid way to ensure Markov abstract states. However, as we discuss in Sec. \ref{sec:bg-pixel-prediction}, the pixel prediction objective is misaligned with the basic goal of state abstraction, since it must effectively throw away no information. It is clear from Figures \ref{fig:rep-vis-tall} and \ref{fig:rep-eval-30k} that our method is able to reliably learn a Markov representation about ten times faster than pixel prediction, which reflects the fact that the latter is a fundamentally more challenging objective.

\newpage

\section{DeepMind Control Experiment with RBF-DQN}
\label{appendix:rbfdqn}
Recently, \citet{asadi2021deep} showed how to use radial basis functions for value-function based RL in problems with continuous action spaces. When trained with ground-truth state information, RBF-DQN achieved state-of-the-art performance on several continuous control tasks; however, to our knowledge, the algorithm has not yet been used for image-based domains.

\begin{figure}[!htbp]
  \centering
  \includegraphics[width=0.7\linewidth]{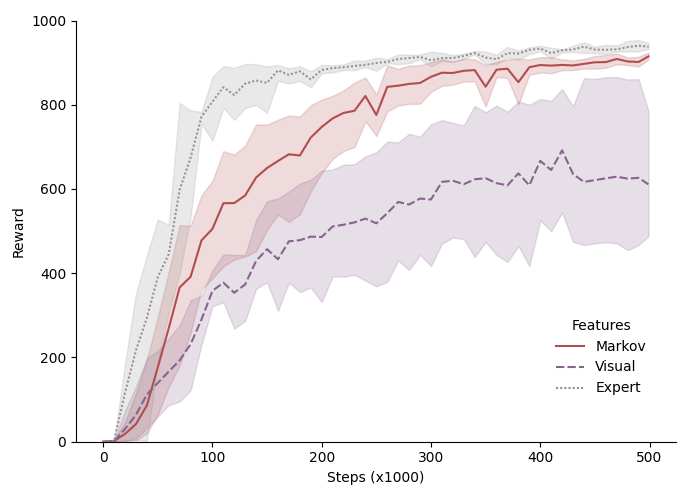}
  \captionof{figure}{Mean episode reward for RBF-DQN on ``Finger, Spin'' with Markov, visual, and expert features. Adding the Markov objective dramatically improves performance over the visual baseline. (Markov -- 5 seeds; Visual -- 6 seeds; Expert -- 3 seeds; shaded regions denote 95\% confidence intervals).}
  \label{fig:rbfdqn}
\end{figure}

We trained RBF-DQN from stacked image inputs on ``Finger, Spin,'' one of the tasks from Section \ref{sec:online},  customizing the authors' PyTorch implementation to add our Markov training objective. We do not use any data augmentation or smoothness loss, and we skip the pretraining phase entirely; we simply add the Markov objective as an auxiliary task during RL. Here we again observe that adding the Markov objective improves learning performance over the visual baseline and approaches the performance of using ground-truth state information (see Figure \ref{fig:rbfdqn}).

\newpage

\section{Investigating the Value of Smoothness in Markov Abstractions}
\label{appendix:smoothness}
Given a Markov abstraction $\phi$, we can always generate another abstraction $\phi'$ by adding a procedural reshuffling of the $\phi$ representation's bits. Since the $\phi'$ representation contains all the information that was in the original representation, $\phi'$ is also a Markov abstraction. However, the new representation may be highly inefficient for learning.

\begin{figure}[!htbp]
  \centering
  \includegraphics[width=0.6\linewidth]{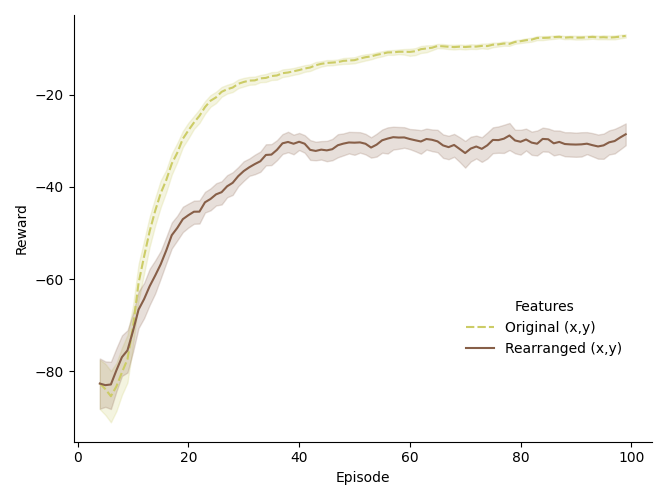}
  \captionof{figure}{Mean episode reward for the $6 \times 6$ gridworld navigation task, comparing original $(x,y)$ position with rearranged $(x,y)$ position. (300 seeds; 5-point moving average; shaded regions denote 95\% confidence intervals).}
  \label{fig:rearranged-xy}
\end{figure}

To demonstrate this, we ran an experiment where we optionally relabeled the positions in the $6 \times 6$ gridworld domain, and trained two agents: one using the smooth, true $(x,y)$ positions, and one using the non-smooth, relabeled positions. Although both representations are Markov, and contain exactly the same information, we observe that the agent trained on the non-smooth positions performed significantly worse (see Figure \ref{fig:rearranged-xy}).

The loss term $\mathcal{L}_{Smooth}$ used in Section \ref{sec:online} penalizes consecutive states that are more than $d_0$ away from each other, thereby encouraging representations to have a high degree of smoothness in addition to being Markov. This approach is similar to the temporal coherence loss proposed by \citet{jonschkowski2015learning}.

\newpage

\section{DeepMind Control Ablation Study}

\begin{figure}[!htbp]
  \centering
  \includegraphics[width=\linewidth]{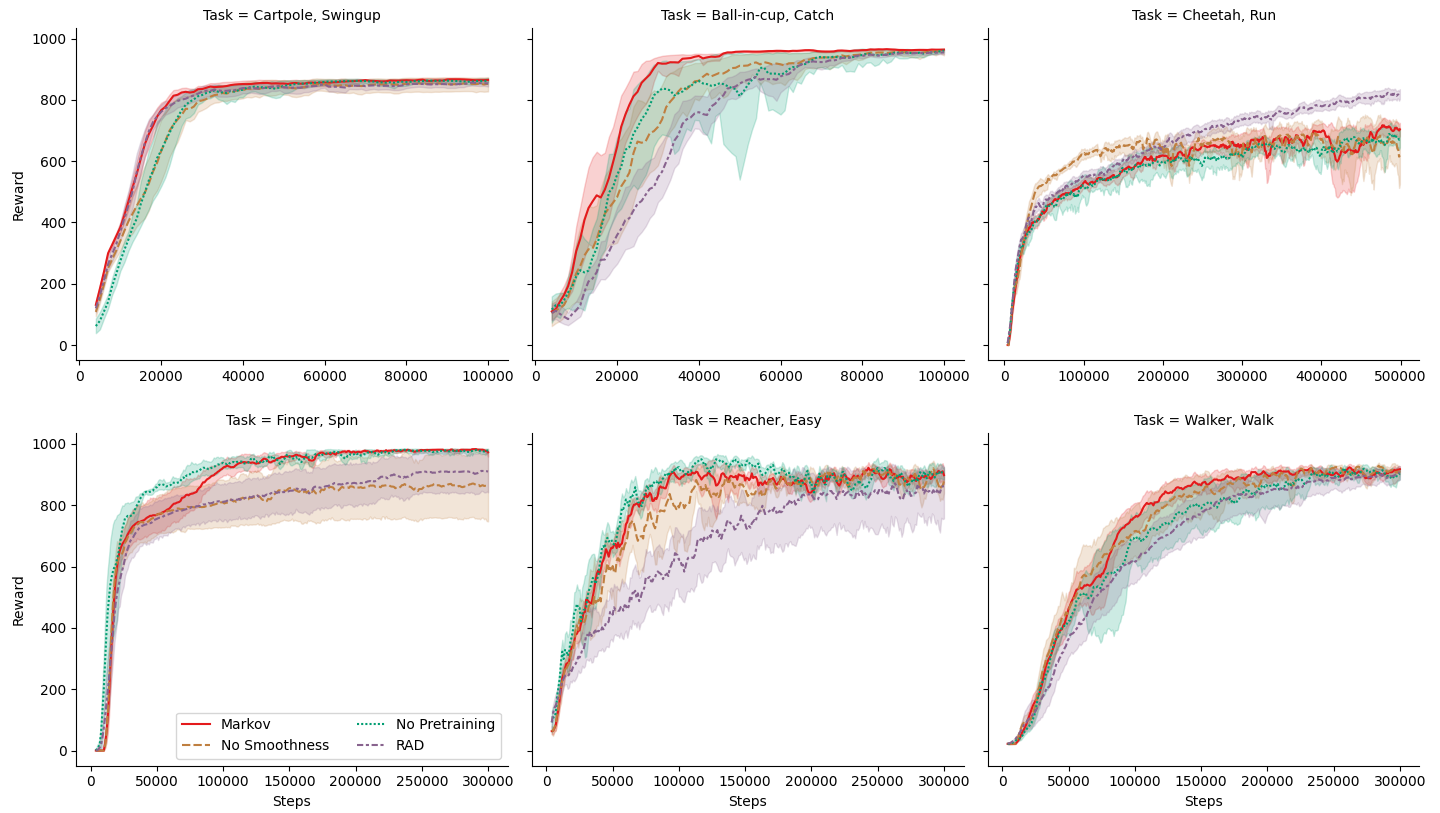}
  \caption{Ablation results for DeepMind Control Suite. Each plot shows mean episode reward vs. environment steps. (Markov and RAD -- 10 seeds; others -- 6 seeds; 5-point moving average; shaded regions denote 90\% confidence intervals).}
  \label{fig:ablation}
\end{figure}

We ran an ablation study to evaluate which aspects of our training objective were most beneficial for the DeepMind Control domains. We considered the RAD implementation and its Markov variant from Section \ref{sec:online}, as well as modifications to the Markov objective that removed either the pretraining phase or the smoothness loss, $\mathcal{L}_{Smooth}$ (see Figure \ref{fig:ablation}).

Overall, the ablations perform slightly worse than the original Markov objective. Both ablations still have better performance than RAD on three of six domains, but are tied or slightly worse on the others. Interestingly, removing pretraining actually results in a slight \emph{improvement} over Markov on Finger. Removing smoothness tends to degrade performance, although, for Cheetah, it leads to the fastest initial learning phase of any method.

We suspect the results on Cheetah are worse than the RAD baseline because the experiences used to learn the representations do not cover enough of the state space. Learning then slows down as the agent starts to see more states from outside of those used to train its current representation. As we point out in Section \ref{sec:bg-smoothness}, it can be helpful to incorporate a more sophisticated exploration strategy to overcome these sorts of state coverage issues. Our approach is agnostic to the choice of exploration algorithm, and we see this as an important direction for future work.

\end{document}